\documentclass[sn-mathphys,iicol,Numbered]{sn-jnl}

\usepackage{lineno}
\usepackage{graphicx}%
\usepackage{multirow}%
\usepackage{amsmath,amssymb,amsfonts}%
\usepackage{amsthm}%
\usepackage{mathrsfs}%
\usepackage[title]{appendix}%
\usepackage{xcolor}%
\usepackage{textcomp}%
\usepackage{manyfoot}%
\usepackage{booktabs}%
\usepackage{algorithm}%
\usepackage{algorithmicx}%
\usepackage{algpseudocode}%
\usepackage{listings}%
\usepackage[capitalize]{cleveref}
\usepackage{bm}
\usepackage{caption}
\usepackage{framed}  
\usepackage{subcaption}

\usepackage[font=small,skip=0pt]{caption}
\usepackage{hyperref}
\hypersetup{
    colorlinks,
    citecolor=blue,
    filecolor=black,
    linkcolor=black,
    urlcolor=black,
}

\usepackage{mathtools}
\usepackage{units}

\DeclarePairedDelimiterXPP\BigOSI[2]%
  {\mathcal{O}}{(}{)}{}%
  {\SI{#1}{#2}}

\usepackage{subfiles}
\begin{document}
\begin{titlepage}
\title{
Learned 3D volumetric recovery of clouds and its uncertainty for climate analysis
}
\end{titlepage}


\author[1]{\fnm{Roi} \sur{Ronen}}\email{ronen.roi@gmail.com}


\author[2]{\fnm{Ilan} \sur{Koren}}

\author[3]{\fnm{Aviad} \sur{Levis}}

\author[4,5]{\fnm{Eshkol} \sur{Eytan}}

\author[1]{\fnm{Vadim} \sur{Holodovsky}}

\author[1]{\fnm{Yoav~Y.} \sur{Schechner}}\email{yoav@ee.technion.ac.il}

\affil[1]{\orgdiv{Viterbi Faculty
of Electrical \& Computer Engineering}, \orgname{Technion - Israel Institute of Technology}, \orgaddress{\street{Technion city}, \city{Haifa}, \postcode{3200003}, \country{Israel}}}

\affil[2]{\orgdiv{Depart. of Earth \& Planetary Sciences}, \orgname{Weizmann Institute of Science}, \orgaddress{\street{Herzl St 234}, \city{Rehovot}, \postcode{7610001}, \country{Israel}}}

\affil[3]{\orgdiv{Depart. of Computing \& Mathematical Sciences}, \orgname{California Institute of Technology}, \orgaddress{\street{1200 E California Blvd}, \city{Pasadena}, \postcode{91125}, \state{CA}, \country{USA}}}

\affil[4]{\orgdiv{Chemical Sciences Laboratory}, \orgname{National Oceanic and Atmospheric Administration}, \orgaddress{\street{325 Broadway}, \city{Boulder}, \postcode{80305}, \state{CO}, \country{USA}}}

\affil[5]{\orgdiv{Cooperative Institute for Research in Environmental Sciences}, \orgname{University of Colorado Boulder}, \orgaddress{\street{1665 Central Campus Mall}, \city{Boulder}, \postcode{80309}, \state{CO}, \country{USA}}}

\abstract{

Significant uncertainty in climate prediction and cloud physics is tied to observational gaps relating to shallow scattered clouds. Addressing these challenges requires remote sensing of their three-dimensional (3D) heterogeneous volumetric scattering content. This calls for passive scattering computed tomography (CT).  We design a learning-based model (ProbCT) to achieve CT of such clouds, based on noisy multi-view spaceborne images. ProbCT infers  – for the first time – the posterior probability distribution of the heterogeneous extinction coefficient, per 3D location. This yields arbitrary valuable statistics, e.g., the 3D field of the most probable extinction and its uncertainty.  ProbCT uses a neural-field representation, making essentially real-time inference. ProbCT undergoes supervised training by a new labeled multi-class database of physics-based volumetric fields of clouds and their corresponding images.
To improve out-of-distribution inference, we incorporate self-supervised learning through differential rendering.
We demonstrate the approach in simulations and on real-world data, and indicate the relevance of 3D recovery and uncertainty to precipitation and renewable energy.

}


\maketitle

\section{Introduction}
\label{sec1}

\begin{figure*}[t]
 \centering
\includegraphics[width=1.0\linewidth]{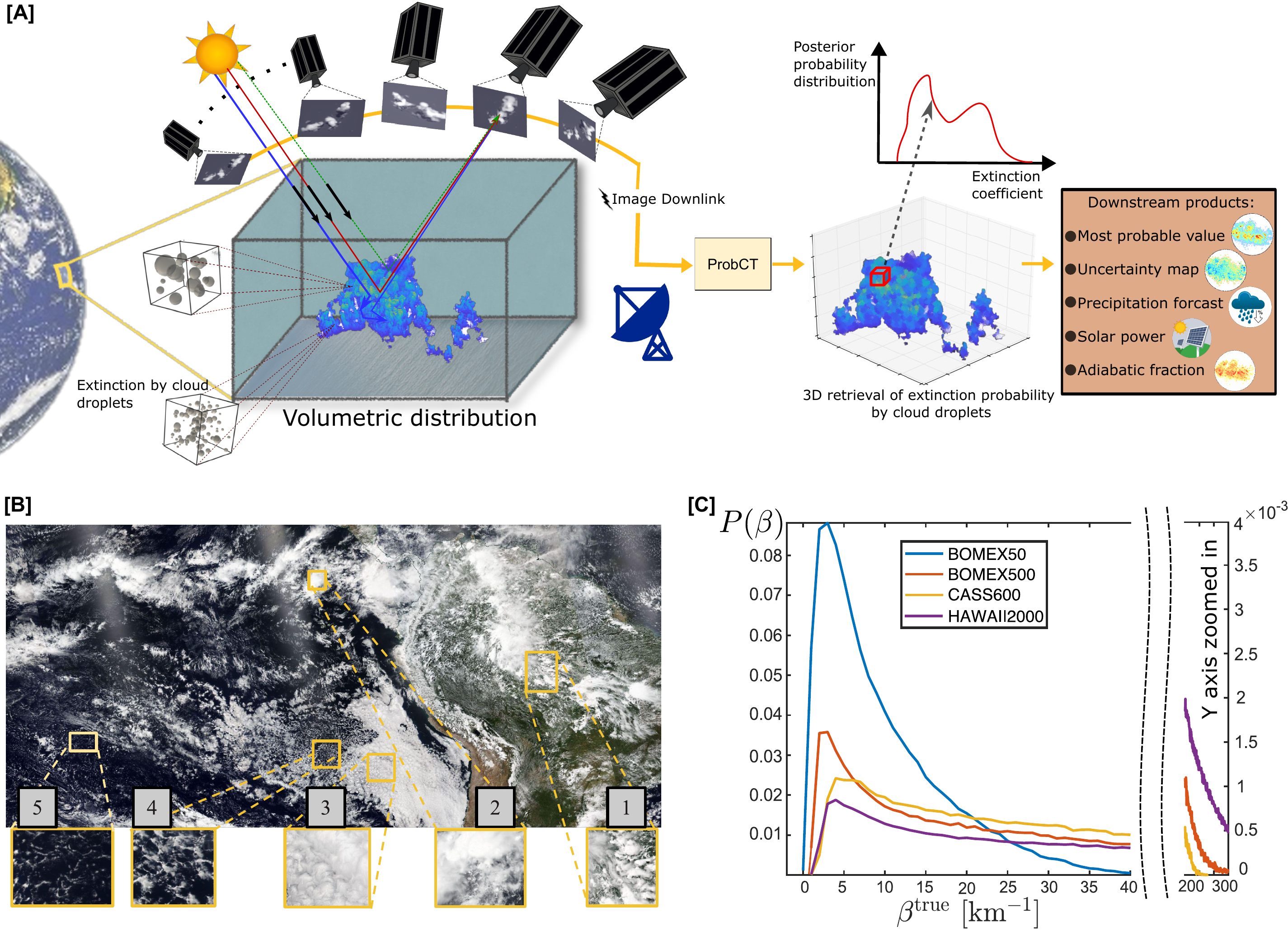}
  \caption{ 
  [A] Images are captured by a coordinated satellite formation (as CloudCT). The images are processed by ProbCT. ProbCT infers the posterior probability distribution of the cloud extinction coefficient at any point in a  3D domain. It thus infers a volumetric map of probability distributions (a distribution per location).
   This facilitates several product, such as: 3D maps of the most probable value and uncertainty of the cloud extinction coefficient, precipitation forecast or uncertainty of solar power at ground level.  
  [B] 
    Several classes of clouds,  imaged by the VIIRS instrument onboard the NOAA-20 satellite:
    \{1\} Anvils of continental deep convective systems; \{2\} 
    Marine deep convective systems;
   Marine stratocumulus deck with closed \{3\} and open \{4\} cells; \{5\} Trade cumulus. 
   [C] Statistics of four classes of simulated cumulus clouds, based on empirical environmental boundary conditions. 
  }
  \label{fig:simulated_clouds}
\end{figure*}

Clouds play a key role in the climate system by modulating  incoming and outgoing radiation energy~\cite{trenberth_2009}. They are controlled by complex thermodynamic, microphysical and radiative processes with all-coupled feedback. Clouds have various types (classes), with examples shown in Fig.~\ref{fig:simulated_clouds}. 

Out of all cloud types, shallow scattered clouds pose the largest challenges~\cite{zelinka_2020,IPCC_AR6_CH7} to cloud and climate prediction. Shallow clouds are regarded as the main coolers of the climate system, thus a highly important question is their feedback on global warming: positive feedback would mean acceleration of greenhouse warming~\cite{IPCC_AR6_CH7}. Moreover, these clouds are not resolved by global climate models, but are represented by sub-grid parameters that are supposed to capture their overall effect and sensitivities~\cite{neggers_2013}. These clouds and their properties are thus recognized as one of the largest sources of climate prediction uncertainty~\cite{bony_2005,ceppi2021observational}. 

These clouds are theoretically challenging to cloud physics, being heterogeneous and sensitive to mixing with their dry three dimensional (3D) environments. Turbulent mixing is complex and challenged by lack of dense 3D measurements \cite{khain_book,derooy_2013,gerber2000structure}. Mixing (thus dilution) in clouds is quantified by the {\em adiabatic fraction} (AF)~\cite{,grosvenor2018remote,lu2023temperature}. It is desirable to map the AF in 3D on a global scale.

To meet needs of prediction and cloud physics requires much better observations, while introducing a remote-sensing challenge. Current analyses in remote sensing~\cite{nakajima1990determination}  assume a {\em plane parallel} structure, where a cloud is very large and homogeneous in broad areas: then, radiative transfer (RT) is mainly modeled vertically. This simplified assumption enabled sensing when computers were weak, satellites were very expensive, and before machine-learning advanced sufficiently. However, this assumption breaks down, leading to significant errors, in shallow scattered clouds: there, RT and heterogeneity vary significantly in 3D.  

Future observations will thus require recovery of volumetric cloud content in 3D, using deca-meter image resolution. Moreover, for scientific use, the {\em uncertainty} of results must also be reported. 

This paper leverages advances in space engineering and machine learning, to help close the sensing and analysis gap in a non-traditional approach.
The advent of nano-satellites lowers costs sufficiently, to make it feasible to create constellations of many satellites. Moreover, 
multiple satellites can fly in a coordinated {\em formation}. From orbit, the satellites can point to the same region, to {\em simultaneously} image any cloud field from multiple directions.  This concept now progresses towards demonstration by the CloudCT space mission~\cite{schilling2019cloudct,tzabari2021cloudct}, funded by the ERC. As illustrated in Fig.~\ref{fig:simulated_clouds}A.


Multi-view image measurements are suited for computed tomography (CT) of volumetric domains. However, cloud imaging is passive, relying solely on sunlight that is {\em multiply scattered} by droplets and other scene components. Scattering {\em creates} the signal. Raw image data relates to 3D volumetric cloud structure by 3D RT, which is the basis of image rendering. CT in this context is an inverse scattering problem, that may iteratively leverage differentiable rendering. 

This concept brings new challenges: (a) Numerically, 3D~RT is a nonlinear recursive operation. It is computationally very difficult to invert, even by physics-based differential rendering. 
For large scale, it is impractical to use
differential rendering in an iterative optimization approach~\cite{levis2020multi,levis2015airborne,aides2020distributed,veikherman2014clouds,tzabari2022settings,fielding2014novel}, for 3D scattering-based CT. For example, optimization by
physics-based differential rendering takes about 20 minutes~\cite{levis2015airborne,sde20213deepct,ronen2022variable} per square kilometer at 50 meter resolution, which is about 3 orders of magnitude slower than the average downlink time from space for a CloudCT formation.  
(b) Nonlinearity challenges computation of {\em uncertainty}, which is very important for to science and downstream technological applications.  
(c) Contrary to controlled imaging settings as in microscopy and medical CT, spaceborne imaging has variable geometry, due to the motion of the cooperating platforms.

We address these challenges by machine learning. Machine learning shifts the computational burden to a {\em training stage}. Consequently, at inference, large data can be scalably analyzed at rates expected from spaceborne remote sensing. The key facilitating technologies mainly include {\em neural fields}~\cite{yu2021pixelnerf,verbin2022ref,liu2022recovery} and self-supervised learning for domain adaptation. In addition, we make use of an encoder-decoder architecture, differential rendering and parallelism enabled in hardware by graphical processing units (GPUs). 

\begin{figure*}[t]
 \centering
  \includegraphics[width=1\linewidth]{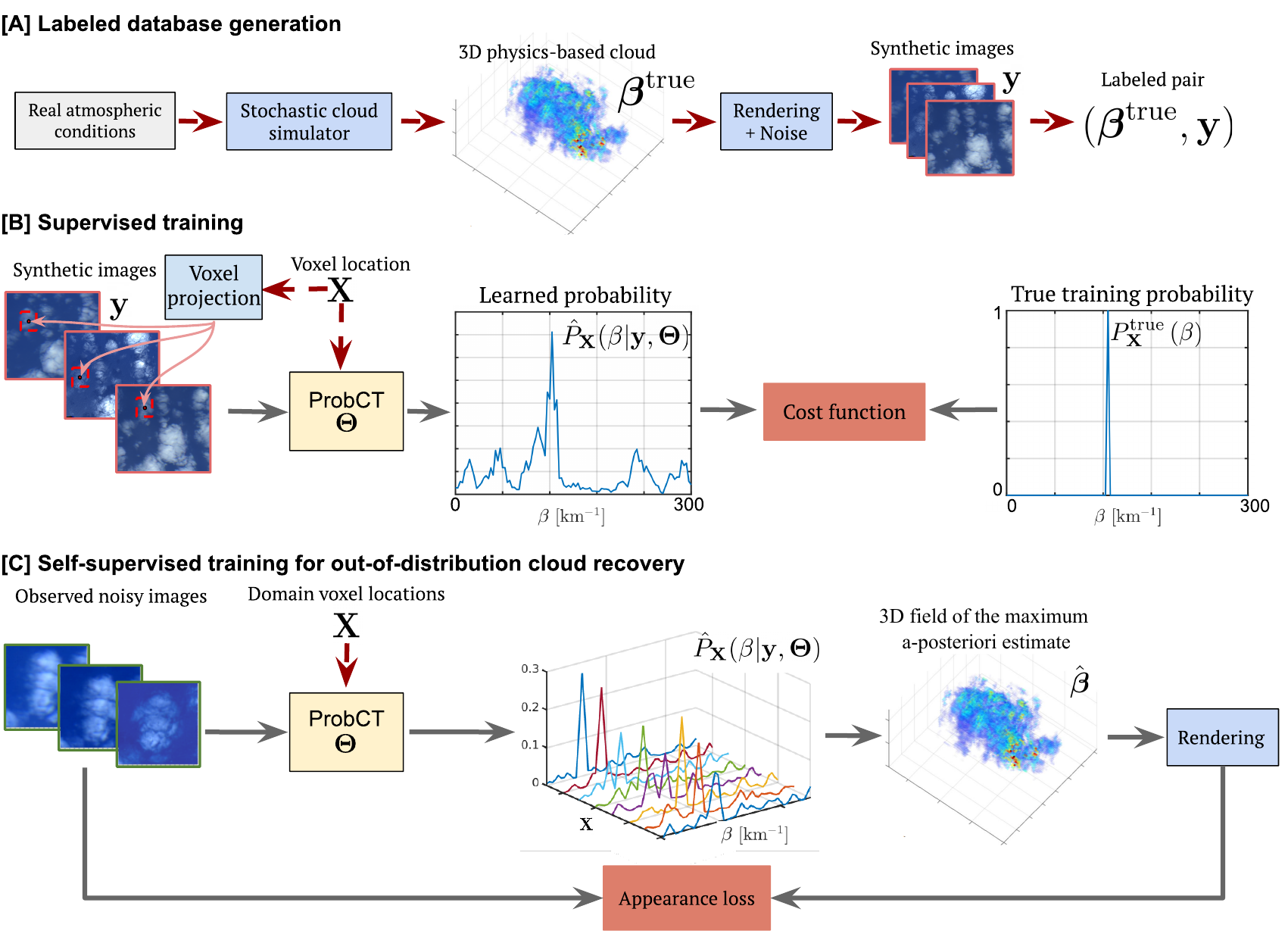}
  \caption{[A] A physics-based cloud simulator generates a random 3D cloud fields ${\boldsymbol \beta}^{\rm true}$. The scene is then physically rendered to yield corresponding multi-view images ${\bf y}$. 
 [B] Supervised volumetric training of ProbCT (tuning the DNN parameters, $\boldsymbol{\Theta}$). 
 ProbCT trains on pairs of labeled data $({\boldsymbol \beta}^{\rm true},{\bf y})$ to  estimate a posterior probability distribution, ${\hat P}_{{\bf X}}(\beta|{\bf y},{\boldsymbol \Theta})$, per 3D  location ${\bf X}$. 
  This probability  is compared to the true probability at this location, 
  ${ P}^{\rm true}_{{\bf X}}(\beta)$, which is a delta function. The comparison uses the Kullback-Leibler divergence.
  [C] Self-supervised training, to improve inference of OOD clouds. Given observed cloud images, ProbCT estimates   
  ${\hat P}_{{\bf X}}(\beta|{\bf y},{\boldsymbol \Theta})$, yielding
  a maximum a-posteriori  estimate  ${\hat {\boldsymbol \beta}}$. Using 
  ${\hat {\boldsymbol \beta}}$ in RT yields re-projected rendered images.
  These images are compared to the observed images, to update ${\boldsymbol \Theta}$.
  }
  \label{fig:ProbCT_train1}
\end{figure*}

Our pipeline is shown in \cref{fig:ProbCT_train1}.
A model trained on clouds implicitly learns and encodes the structural nature  of clouds. This serves as a prior during cloud inference. 
Training a model would ideally use a lot of empirical labeled data. If it were possible, that would mean a large set of ground-truth clouds in nature, whose volumetric contents are known in 3D, by hundreds of thousands of in-situ simultaneous measurements per cloud. It is not feasible to obtain such in-situ large data. Our alternative approach to obtaining ground-truth clouds leverages dynamical models of clouds, created and validated by the cloud physics community. They integrate thermodynamics, fluid dynamics,  condensation and evaporation dynamics at high spatio-temporal resolution, based on empiric  environmental boundary conditions.
Nature has different classes of clouds, each class emerging from a different type of environment.
For this reason, simulations use a variety of environmental conditions, to yield different cloud classes, each class having its own statistics.

The resulting random clouds are physically sound, but their statistics are distributed according to the sets of boundary conditions. If only such clouds are used for training, the trained model may be vulnerable to out-of-distribution (OOD) inputs at inference. Therefore, learning-based scattering CT of clouds should handle domain adaptation. Inspired by physics-based learning~\cite{kadambi2023incorporating}, after supervised training by labeled simulated data, we employ self-supervised learning  using real-world empirically acquired cloud images. This helps inference of OOD samples.

In this work, we introduce  the following contributions:\\
(1) A deep neural network (DNN) model, termed {\em ProbCT}. Its input comprises multi-view spaceborne noisy images taken from a formation of satellites that can be in a variable geometry. The input can express non-rigid orbital geometry and noise that naturally exists in images of relevant sensors. At inference, ProbCT outputs per location a {\em function}: the {\em posterior probability distribution} of the extinction coefficient. Consequently, per location, it is possible to estimate not only an optimal value of the extinction coefficient, but also the {\em uncertainty} of the estimation. Inference run-time by ProbCT is comparable to the downlink rate from orbit.\\
(2) An expansive multi-class labeled database for supervised learning of shallow small clouds and testing of OOD scenes. \\
(3) A self-supervised training technique for improving inference of OOD cloud fields. \\
(4) Examples of downstream tasks affected by the inferred 3D structure and the quantified uncertainty.

\begin{table*}[ht]

\begin{subtable}{1\textwidth}
\centering
   \begin{tabular}{lrrcccc}
\toprule
Dataset    & \#Train & \#Test & Voxel size {[}m{]}      & Grid size      & Camera & Pixel footprint        \\
\midrule
BOMEX50    & 1660    & 203    & $50\times 50 \times 40$ & $32\times 32 \times 64$ & perspective & 20[m]  \\
BOMEX500   & 6001    & 566    & $50\times 50 \times 40$ & $32\times 32 \times 32$ & perspective & 20[m]   \\
CASS600    & 10908   & 1000   & $50\times 50 \times 40$ & $64\times 64 \times 32$ & perspective & 20[m]   \\
HAWAII2000 & 1227    & 722    & $50\times 50 \times 20$ & $32\times 32 \times 64$ & perspective & 20[m]  \\
BOMEX500-Aux & 4418   & $\times$    & $50\times 50 \times 40$ & $32\times 32 \times 32$ & pushbroom & 10[m]    \\
\botrule
\end{tabular}%
   \caption{}\label{tab:datasets}
\end{subtable}

\begin{subtable}{1\textwidth}
\centering

    \medskip

\begin{tabular}{@{}lccccc@{}}
\toprule
Train \textbackslash Test & BOMEX50      & BOMEX500   & CASS600      & HAWAII2000 &   \\ 
\midrule
BOMEX50                   & 0.33$\pm$0.11   & $\times$ & $\times$            & $\times$          &    \\
BOMEX500                  & 0.49$\pm$0.10    & 0.33$\pm$0.13 & 0.36$\pm$0.14  & 0.52$\pm$0.20  &   \\
CASS600                   & $\times$            & 0.65$\pm$0.13 & 0.22$\pm$0.06 & $\times$          &  \\
HAWAII2000                & $\times$            &$\times$          & $\times$            & 0.44$\pm$0.17 & \\

\midrule
Physics-based~\cite{levis2015airborne}             & 0.51$\pm$0.07 & 0.97$\pm$0.24 & 0.76$\pm$0.38   & 0.82$\pm$0.27 & \\
\botrule
\end{tabular}
   \caption{}\label{tab:results}
\end{subtable}
\caption{(a) Specifications of simulated cloud datasets. Each has a different number of training and testing examples, voxel size resolution and domain size. This requires flexibility of the analysis system. The BOMEX500 and  BOMEX500-Aux sets include perturbations to the imaging geometry.
(b) Results of cloud volumetric recovery, measured by  the  mean and standard deviation of 
$\epsilon$ (Eq.~\ref{eq:erros}). Each row represents the training class of clouds from which   examples  are drawn for supervised training. Columns refer to the test class from which clouds are drawn. 
The main diagonal summarizes ID inference errors. Off-diagonals apply to OOD inference. 
ProbCT  outperforms an existing  physics-based solver~\cite{levis2015airborne} for both ID and OOD tests across all datasets. Moreover, ProbCT inference  is about $\times 1000$ faster than the physics-based solver.
} \label{tab:three_tables}
\end{table*}

\section{Results}
\label{sec:results}

\subsection{Expansive Database}
\label{sec:database}

Validated dynamical system simulators form synthetic, physics-based clouds. The simulators rely on initial atmospheric conditions, including aerosols. There are well-studied initial atmospheric conditions, termed BOMEX, CASS and HAWAII. Based on them, we have derived several classes  (thus sets) of simulated cloud fields. In this work, to enable OOD studies, we derive datasets 
BOMEX50 and HAWAII2000, expanding smaller data~\cite{sde20213deepct} in sets termed BOMEX500 and CASS600. The suffix number is the aerosol concentration in ${\rm particles/{\rm cm}^{3}}$.
We also use an augmentation using the BOMEX500-Aux dataset: there, for each cloud of the BOMEX500 set, the liquid water content (LWC) is multiplied by 1/10. Some statistics of these classes appear in Table~\ref{tab:datasets}  and \cref{fig:simulated_clouds}[C].

Each class of clouds has a different statistical distribution. The cloud sets have different spatial textures, spanning different domain and voxel sizes.
Per class, a simulation results in a labeled ground-truth volumetric scene. A scene is characterized by a spatially-varying field of the optical extinction coefficient. A vector 
$ {\boldsymbol \beta}^{\rm true}$ represents this coefficient
in a grid of  voxels.  
Each such scene can be converted to corresponding image data (denoted ${\bf y}$) at any optical wavelength and viewpoint poses, using RT.  The cloud scenes and the corresponding images constitute labeled databases. 

We are motivated by the CloudCT~\cite{schilling2019cloudct,tzabari2021cloudct}, 
formation: $10$ nano-satellites, having 100km between nearest neighboring, perspective viewpoints, orbiting
500km high. 
So, we use a viewing geometry similar to that, as illustrated in Fig.~\ref{fig:simulated_clouds}A. Then, the solar zenith angle is $25^\circ$. 
We use a spectral band around \unit[672]{nm} and random image noise, whose specifications are typical to the CloudCT payload (see \cref{sec:method}). 
Rendered images have $116\times116$ pixels, with \unit[20]{m/pixel} at nadir.

Additionally, we use rendered images corresponding to NASA's AirMSPI instrument. It takes 9  pushbroom multi-angular images in a 
$\pm67^\circ$ angular span along the flight path at \unit[20]{km} altitude, with \unit[10]{m} resolution at nadir, around wavelength $660[{\rm nm}]$. 
Redenring here uses the synthetic  cloud scenes from the BOMEX500-Aux dataset using five different AirMSPI flight experiments having random perturbations of viewpoints.
The rendered AirMSPI training images use random radiometric noise, having the specifications~\cite{ASDC2013AirMSPI} of this sensor.

\subsection{Simulated Tests}
\label{sec:sim}

\begin{figure*}[!htp]
 \centering
  \includegraphics[width=0.95\linewidth]{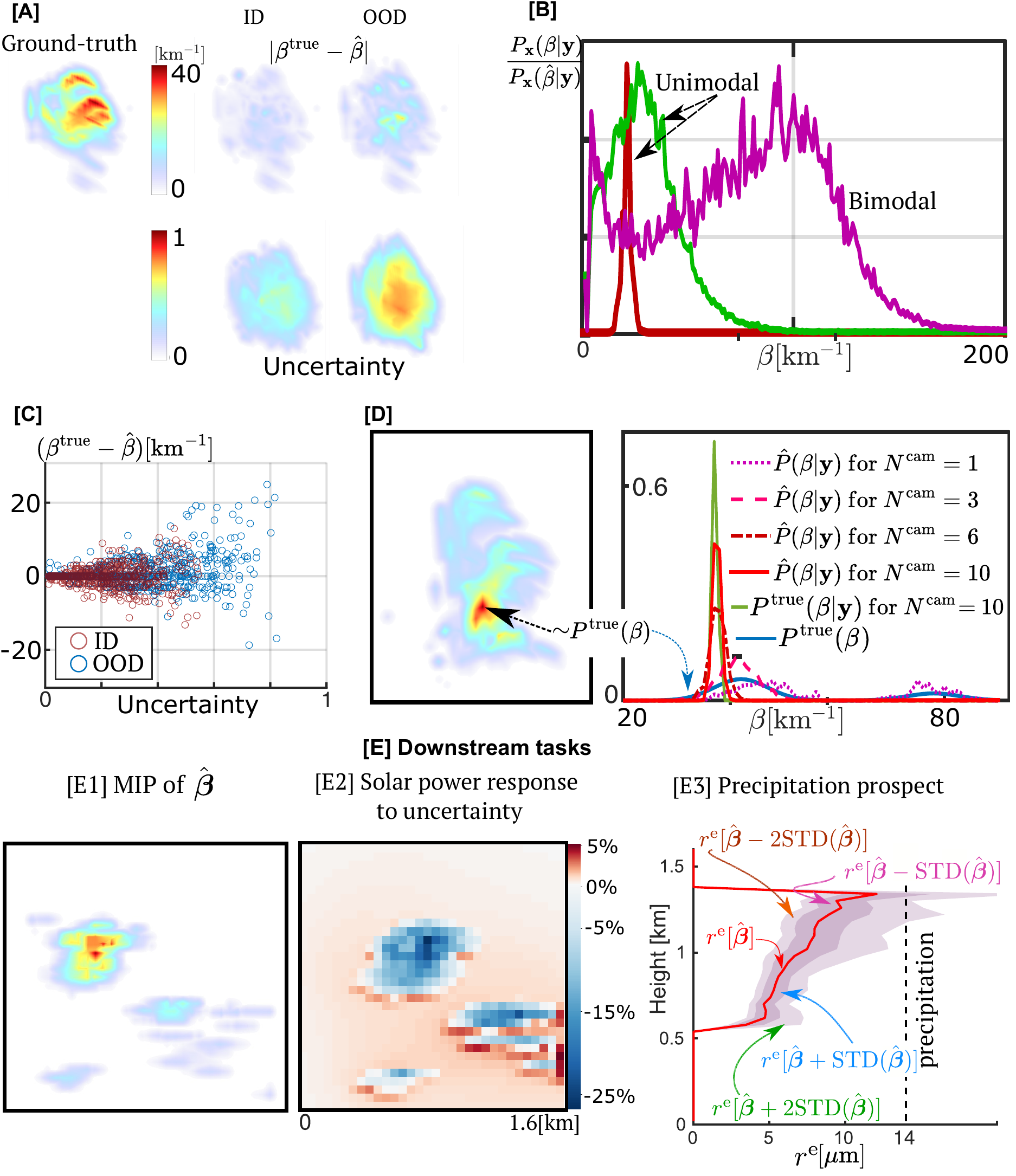}
  \caption{\textbf{[A]} Visualizations of 3D volumetric fields by MIP at $45^\circ$ off-nadir: A labeled BOMEX50 test cloud, its estimation error and uncertainty (normalized entropy), which increases at the cloud core.
  \textbf{[B]}
  Sample inferred probability distributions from a cloud shown in 
  [A] (an OOD test), normalized by
  the MAP value. 
  \textbf{[C]}
  Inferred results at 2000 voxels, randomly sampled across the BOMEX50 test set.  
 A high inferred normalized entropy (uncertainty) indeed implies a possible large absolute error.
 Large errors of $\hat \beta$ rarely occur when the inferred entropy is low.  
   \textbf{[D]}
Learning the posterior probability distribution  of clouds that
differ by a single voxel (pointed out by a black arrow).
The cloud is visualized by MIP at $45^\circ$ off-nadir. [Right] Blue: the prior probability distribution  from which $\beta$ at the voxel is drawn. Green: the sharply peaked true posterior of $\beta$ in this voxel of a test cloud, by
$N^{\rm cam}=10$. Other lines plot ProbCT inferences for different number of views $N^{\rm cam}$. 
   [E1] A recovered BOMEX500 test scene. The 
uncertainty in ${\hat {\boldsymbol\beta}}$ propagates to downstream forecasting tasks: [E2]
renewable solar power generation on the ground~\cite{gristey2020surface}
(see Eq.~\ref{eq:rel_responce})
and [E3] droplet effective radius \mbox{$r^{\rm e}$}.
A value \mbox{$r^{\rm e}>14\mu{\rm m}$} is a {\em precipitation trigger}~\cite{rosenfeld1994retrieving}, yielding rain and dramatic shortening of cloud life.
  }
  \label{fig:3d_bomex}
\end{figure*}

Per location {\bf X}, ProbCT estimates a {\em function}: the posterior probability distribution of the extinction coefficient $\beta$, that is ${\hat P}_{{\bf X}}(\beta|{\bf y})$. A probability distribution enables extraction of statistics. These include, per-location, the maximum a-posteriori (MAP) value ${\hat \beta}$, an expected value, and uncertainty measures such as  standard deviation (STD) or normalized entropy $0\leq H_{\bf X}^{\rm norm}\leq 1$. We illustrate this in an example. 
\cref{fig:3d_bomex}[A] visualizes a cloud, using maximum intensity projection (MIP). 
Based on noisy multi-view images of the cloud, ProbCT infers ${\hat P}_{{\bf X}}(\beta|{\bf y})$. For some locations in the cloud, example probability distribution functions  inferred by ProbCT are plotted in \cref{fig:3d_bomex}[B].
3D maps of the error $|\hat {\boldsymbol \beta}- {\boldsymbol \beta}^{\rm true}|$ of the MAP estimator
and the corresponding  $H_{\bf X}^{\rm norm}$ are visualized by MIP in \cref{fig:3d_bomex}[A]. 
As expected, the uncertainty tends to increase with the estimation error.

This example points to an additional issue. At testing, an inferred scene might belong to a type of clouds used in supervised training, i.e., meeting in-distribution (ID) inference. In general, however, it is expected that an inferred scene would have clouds from an unknown type. Thus generally, at inference, a cloud may not belong to a type (class) used in supervised training. The cloud is then out of the training distribution (OOD). Then indeed, in the example of \cref{fig:3d_bomex}[A], both the estimation error and the uncertainty increase when inference is OOD, mainly in the cloud core. 

Over all clouds in the  BOMEX50 test set, we randomly sample voxels with their corresponding MAP estimation. The estimation errors are scatter-plotted vs. $H_{\bf X}^{\rm norm}$ in \cref{fig:3d_bomex}[C]. 
A seen, large errors occur only where the inferred relative entropy  $H_{\bf X}^{\rm norm}$ is high.  That is, ProbCT indicates where its estimation of $\beta$ may fail.

As described in the Methods section (Sec.~\ref{sec:train}), 
ProbCT is trained in two stages, to help it handle OOD clouds. First, ProbCT undergoes supervised training using ground-truth clouds from a database. Afterwards, the ProbCT model is refined by self-supervised training, using only {\em images} whose statistics are similar to images of the inferred clouds. The extinction field $\beta(\bf X)$ can be sampled in a voxel grid, to form a vector ${\boldsymbol \beta}$. To quantitatively evaluate the MAP performance, 
we follow~\cite{levis2015airborne,sde20213deepct,ronen2022variable} and use per scene
\begin{equation}
 \epsilon =\frac {\| { {\boldsymbol \beta}^{\rm true}} - {\hat {\boldsymbol \beta}} \|_1 }
 {\| { {\boldsymbol \beta}^{\rm true}}  \|_1}
  \;.
  \label{eq:erros}
\end{equation}

In \cref{tab:results}, each row represents the dataset from which  3D cloud examples  are drawn for supervised training. 
Therefore, the main diagonal in \cref{tab:results} summarizes ID inference errors, while off-diagonal results in \cref{tab:results} apply to OOD inference.
 For example, the column titled BOMEX50 stands for testing on the BOMEX50 clouds. There, results in the first and second rows  involve supervised training using BOMEX50 or BOMEX500 respectively, followed by refinement using images (not 3D clouds) that correspond to clouds from the BOMEX50 training set.  
 
Over the test sets, whether ID or OOD, ProbCT outperforms 
a physics-based method~\cite{levis2015airborne}, which uses iterative optimization by differential rendering.
Moreover, ProbCT requires on average less than a second per inferred scene, when computation uses a single GPU. In contrast,
recovery using physics-based differential rendering~\cite{levis2015airborne} requires $\sim1000$ seconds. 

The success of ProbCT is not only due to training. Rather, it stems from information carried by multi-view geometry, essential for CT. To show this, we vary the number of viewpoints, $N^{\rm cam}$. The training and testing datasets here are made of clouds that differ by a single voxel in the cloud core.
In this voxel, $\beta$ is sampled randomly from a bimodal probability distribution  $P(\beta)$ (see Sec.~\ref{sec:singlevox}). Images from $N^{\rm cam}$ viewpoints are rendered, compounded with noise, yielding image data $\bf y$. 
The data likelihood $P({\bf y}|\beta)$ is set by the known noise specifications. 
As only a single cloud voxel is unknown, the true posterior of a test cloud can be calculated by Bayes rule: 
\begin{equation}
 P^{\rm true}(\beta|{\bf y})=[P({\bf y}|\beta)P(\beta)]/
  \left[
     \int  P({\bf y}|\beta)P(\beta)d\beta
  \right]   
  \;.
  \label{eq:bayes}
\end{equation}
Per $N^{\rm cam}$, a ProbCT model is trained; then the corresponding trained ProbCT model infers ${\hat P}(\beta|{\bf y})$ in that voxel. 
Results are presented in \cref{fig:3d_bomex}[D].
  At \mbox{$N^{\rm cam}=1$}  data is insufficient for CT recovery, despite training, and indeed inference  yields $\hat P(\beta|{\bf y})\rightarrow P^{\rm true}(\beta|{\bf y})\sim P^{\rm true}(\beta)$.
On the other hand, as $N^{\rm cam}$ increases, $\hat P(\beta|{\bf y})\rightarrow P^{\rm true}(\beta|{\bf y})$, which is sharply peaked at ${\beta}^{\rm true}$.
An additional interpretable example  is shown in the supplementary material. 







\cref{fig:3d_bomex}[E] demonstrates that inferred uncertainty is important for downstream tasks, specifically, 
forecasts regarding renewable energy and precipitation. 
Renewable solar energy converts light power to electric power by a sequence of processes. 
The Methods section explains how irradiance reaching the ground converts to  electrical current units generated by silicon-based photovoltaic (PV) panels. The current is denoted $i_{\rm PV}({\boldsymbol \beta})$, given scene ${\boldsymbol{\beta}}$.

The atmosphere (including clouds and air molecules) affects  $i_{\rm PV}({\boldsymbol \beta})$ non-linearly in two opposite manners. The reason is that direct solar irradiance is attenuated by the atmosphere, while diffuse irradiance typically increases by atmospheric density via scattering.
We define the relative response of $i_{\rm PV}$ to uncertainty in ${\hat {\boldsymbol \beta}}$ by
\begin{equation}
    i_{\rm PV}^{\rm rel} =  
      \frac{ i_{\rm PV}[{\hat {\boldsymbol \beta}}+{\rm STD}({\hat {\boldsymbol \beta}})] - i_{\rm PV}[{\hat {\boldsymbol \beta}}-{\rm STD}({\hat {\boldsymbol \beta}})]}
     { i_{\rm PV}[{\hat {\boldsymbol \beta}}]}
     \;.
    \label{eq:rel_responce}
\end{equation}

Consider a  $1.6\times1.6${km} field of PV panels, each placed horizontally under a cloud field shown in \cref{fig:3d_bomex}[E1]. 
The corresponding two-dimensional map of  $i_{\rm PV}^{\rm rel}$ is shown in \cref{fig:3d_bomex}[E2].

Precipitation is triggered by a critical size of the cloud droplets~\cite{rosenfeld1994retrieving}. Per voxel, the effective radius of the droplets can be estimated by $\hat{\beta}$ (see \cref{sec:product}).
Let ${r}^{\rm e}[{\boldsymbol{\beta}}]$ be the 
effective radius averaged horizontally at the cloud core.  \cref{fig:3d_bomex}[E3] plots  ${r}^{\rm e}[\hat {\boldsymbol{\beta}}]$ as a function of altitude. The uncertainty in ${\hat {\boldsymbol{\beta}}}$ propagates to
corresponding plots of ${r}^{\rm e}$. 
\cref{fig:3d_bomex}[E3] indicates that while the MAP cloud does not precipitate, the uncertainty in the  estimation points to chances otherwise.


\subsection{AirMSPI real empirical data}
\label{sec:airmspi}

To prepare future space missions for multi-view high resolution cloud imaging, we use (beside rigorous simulations) AirMSPI real-data from NASA. AirMSPI imaging is done sequentially during $\approx 10$ minutes along a flight path. During this time,  clouds drift due to wind. Thus, as a pre-process, we follow~\cite{ronen20214d} to assess and compensate global drift between images. We follow an experiment from~\cite{levis2015airborne,ronen2022variable}. 

\begin{figure*}[t]
 \centering
  \includegraphics[width=1.0\linewidth]{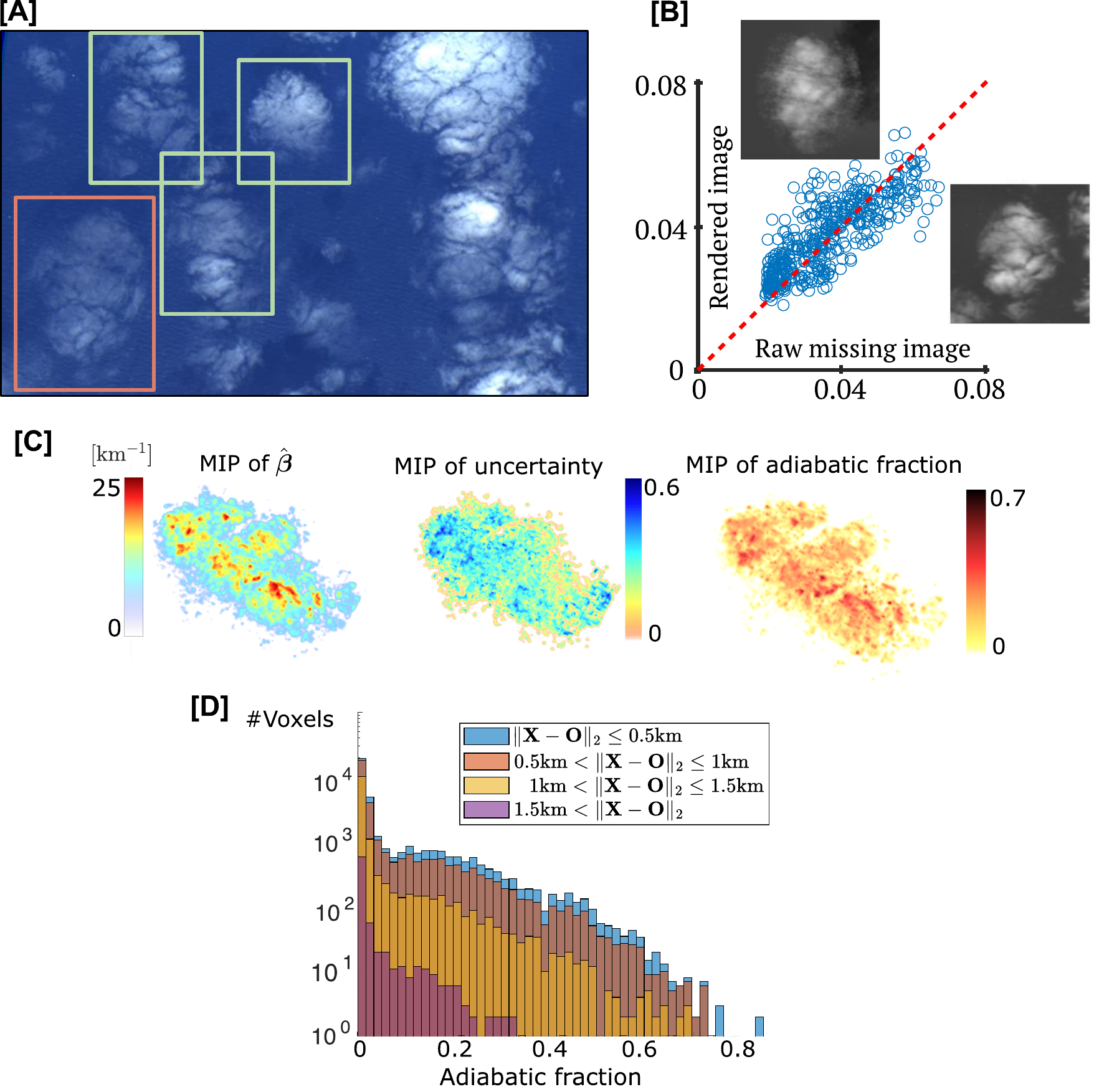}
  \caption{Real-world experiment. [A] Nadir image from a NASA AirMSPI flight at 20:27GMT on February 6, 2010 over 
  32N 123W. 
  Green rectangles: regions used for self-supervised training. Red rectangle: a test domain.
  [B] Comparing (visually and by a scatter plot) an AirMSPI image excluded from inference vs. an image rendered in the corresponding viewpoint, based on
  the inferred cloud.  
   For clarity, the scatter-plot uses $1\%$ of the image pixels.
   [C] MIP of the inferred MAP ${\hat \beta}$ of the cloud, MIP of the uncertainty (normalized entropy), and MIP of an estimated adiabatic fraction (AF).
  [D] Histogram of the estimated AF. Bar colors represent voxel distance from the cloud center $\bf O$.
   As expected, the AF decreases as distance increases from the cloud core. }
  \label{fig:clouds_airmspi}
\end{figure*}

We consider clouds imaged by AirMSPI to be OOD relative to our simulated clouds. So, after supervised learning using simulated labeled clouds, we performed self-supervised learning using real multi-view images observed by AirMSPI. Self-supervised learning uses three clouds of similar characteristics in the field of view. After training, we inferred a fourth volumetric domain (see \cref{fig:clouds_airmspi}), having $72\times 72 \times 32$ voxels, i.e., 165,888 unknowns~\cite{ronen2022variable}.

We check for consistency using cross-validation. For this, we excluded the $+47^\circ$ view from the input: ProbCT then inferred  both ${\hat {\boldsymbol \beta}}$ and the uncertainty field, using only eight viewpoints. Afterwards, we used an RT forward-model 
$ {\cal F}({\hat {\boldsymbol \beta}}) $
to render the missing view. 
Qualitative results of ${\hat {\boldsymbol \beta}}$ are shown in 
\cref{fig:clouds_airmspi}. 
For quantitative results, we use the root mean square error (RMSE) on the excluded image.
The RMSE is measured in $[{\rm Wm^{-2}sr^{-1}nm^{-1}}]$.
ProbCT yields ${\rm RMSE}=1.15$. In comparison, a physics-based solver~\cite{levis2015airborne}, which is not learning-based yields ${\rm RMSE}=1.03$, while running $\approx 3$ orders of magnitude slower.

A novel insight that such recovery can potentially yield is shown for example in \cref{fig:clouds_airmspi}[C,D].  The liquid water content (LWC) around location ${\bf X}$ in a cloud is ${\rm LWC}({\bf X})$  $[\rm g/m^3]$.  On the other hand, by the adiabatic model, LWC is only a function of the altitude $Z$~\cite{acp-eshkol} above the cloud base.
This function, ${\rm LWC^{ad}}({Z})$, can be pre-computed irrespective of scattering CT. The AF is
\begin{equation}
    {\rm AF}({\bf X})=
    \frac{{\rm LWC}({\bf X})}
          {{\rm LWC^{ad}}({Z})}\;.
    \label{eq:AF}
\end{equation}
For a cloud voxel undiluted by mixing with surrounding air, ${\rm AF}=1$. Dilution lowers the AF. To the best of our knowledge, this measure has not been experimentally retrieved in 3D by remote sensing. We present first results in this direction. In this preliminary example, we use a function shown in the supplementary material, drawn from conditions at Barbados. 
The clouds in \cref{fig:clouds_airmspi}[A] are vertically thin geometrically and optically. The results in \cref{fig:clouds_airmspi}[C,D] suggest that near the edges of the cloud (10s of meters) lies a transition zone with ${\rm AF}<0.2$, while the AF continuously increases towards the cloud center. This agrees with theory and observations, that show that near the cloud center typically lies 
an undiluted {\em cloud core}~\cite{eytan2022shallow,konwar2021cloud,gerber1996microphysics}.


\section{Discussion and conclusions}
\label{sec:discusion}

ProbCT is the first system designed to assess a probability distribution per location of a heterogeneous extinction coefficient, across a volumetric scattering domain, and focusing on CT of clouds.
It is trained to accommodate imaging noise, variation in imaging geometry, 
and cloud variability within and across cloud classes. We demonstrate it in ID and OOD tests. 
The ProbCT inference runtime is essentially in real-time concerning data downlink rate from space.
By estimating a posterior probability distribution, ProbCT enables per-location assessment of uncertainty, which
is an essential measure in science.

Fundamentally, inverse scattering is ill-posed: a variety of multiply-scattering volumetric contents can ``explain'' the measured noisy radiance. Hence, the true 
$P(\beta|{\bf y})$ is \emph{not} a delta function, regardless of the estimator of this distribution. Thus, any good estimator (ProbCT as well) is not expected to generally output a delta function for ${\hat P}(\beta|{\bf y})$. Note that here ${\bf y}$ is only image-based. However, there are additional, non-image sources that can be fed into a learning-based system, that enrich ${\bf y}$ and can lower the inferred uncertainty and errors. Specifically, clouds are affected by environmental conditions such as atmospheric temperature and humidity profiles, which are sampled globally by various meteorological instruments~\cite{siebesma2003large}. This can improve the estimation of the {\em veiled core}~\cite{forster2021toward}: this region may comply better with the adiabatic model, which depends on these conditions. The adiabatic model may be implicitly learned by the system.

This work opens the door to new ways of scientific analysis relating to remote sensing and
atmospheric physics. It indicates that complex multi-view images can realistically be acquired
and processed to shed light on hard questions involving 3D heterogeneity and multiple
scattering. Moreover, some principles of ProbCT may be relevant to other domains where multiple scattering and/or reflections are major, such as medical imaging, non-line-of-sight imaging, and reflectometry. 
An extension of this work should aim to recover the joint distribution of several parameters per location (single-scattering albedo, cloud droplet sizes, and their density).  
%

\section{Methods}\label{sec:method}

\subsection{ProbCT Model}
\label{sec:model}

\subsubsection{Architecture}
\label{sec:archi}

\begin{figure*}[!ht]
 \centering
\includegraphics[width=1.0\linewidth]{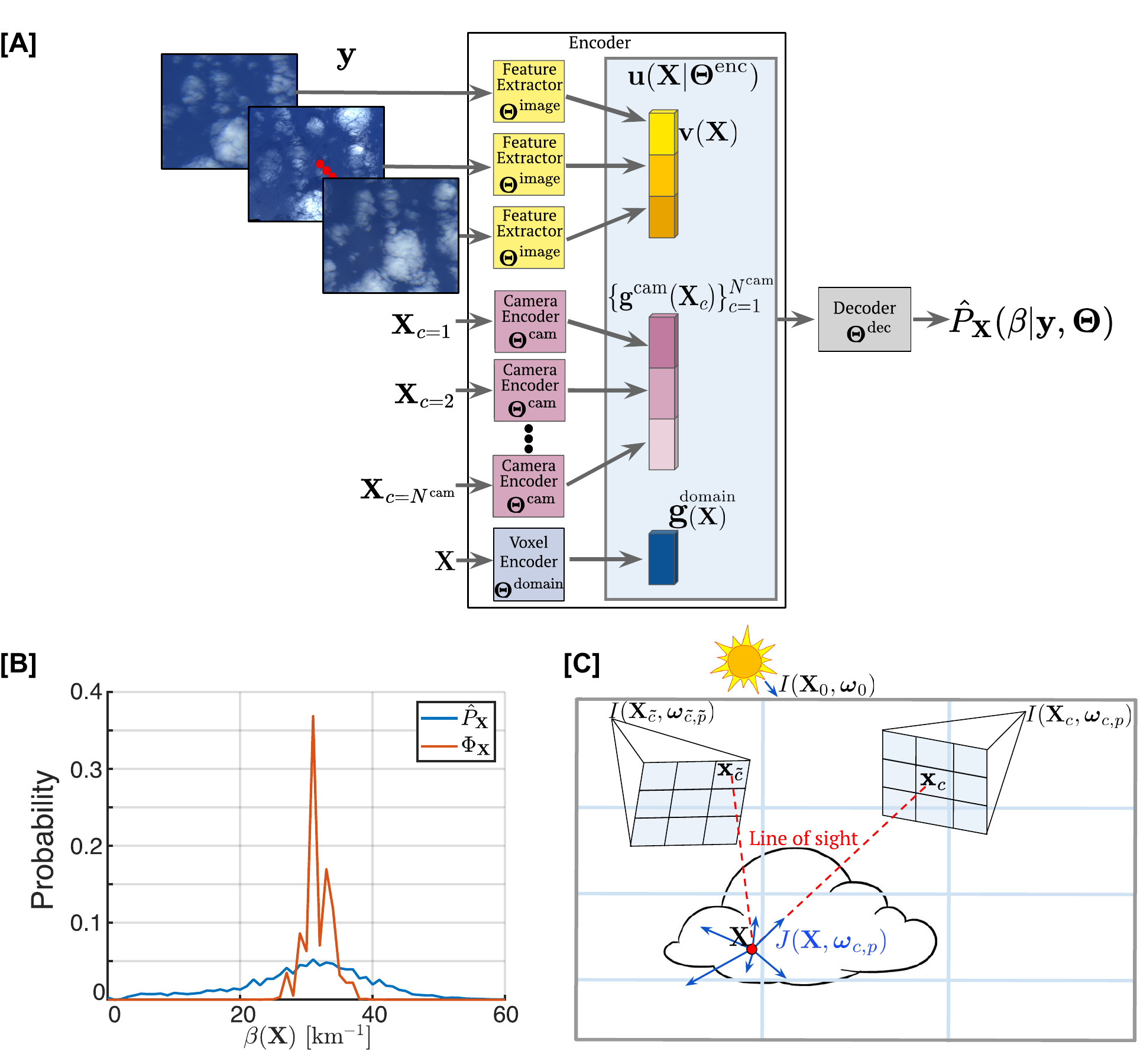}
  \caption{ 
  [A] The ProbCT architecture. 
  A 3D scene is observed from $N^{\rm cam}$ viewpoints, yielding multi-image data ${\bf y}$. All images are processed by the same feature pyramid network, to extract corresponding image feature maps. 
  Per  location ${\bf X}$ the feature maps are sampled at spatial coordinates
  $\{ {\bf x}_c\}_{c=1}^{N^{\rm cam}}$
  corresponding to geometric projections of ${\bf X}$. 
  This leads to a vector ${\bf v}({\bf X})$  of features from all images.
  3D coordinates of the  location ${\bf X}$ and locations $\{{\bf X}_{c}\}_{c=1}^{N^{\rm cam}}$ of the viewpoints are processed using coordinate encoders, resulting respectively in geometric feature  vectors 
  ${\bf g}^{\rm domain}({\bf X})$ and $\{{\bf g}^{\rm cam}({\bf X}_{c})\}_{c=1}^{N^{\rm cam}}$. 
  These vectors are passed to a decoder that infers the posterior probability distribution of the extinction coefficient 
  ${P}_{{\bf X}}(\beta|{\bf y},{\boldsymbol \Theta})$.
  [B] 
  Visualization of the differential ${\tt Smoothmax}$ (Boltzmann)  operator~\cite{asadi2017alternative} in \cref{eq:amp_p}. This form is used during self-supervised training. 
  [C] At boundary point  ${\bf X}_0$, known 
  radiance $I({\bf X}_0, {\boldsymbol \omega}_0)$ is incident in direction ${\boldsymbol \omega}_0$. 
   Radiance  scatters multiple times in the domain. 
  The 3D functions
  $J$ (Eq.~\ref{eq:J}) and the extinction coefficient $\beta$ define the radiance field $I$ by \cref{eq:I}.
  Pixel $p$ of camera $c$ corresponds line of sight to direction ${\boldsymbol \omega}_{c,p}$.
  This pixel samples the radiance $I({\bf X}_{c}, {\boldsymbol \omega}_{c,p})$.
  }
  \label{fig:methods}
\end{figure*}

In this section we describe the architecture of the ProbCT model, presented in \cref{fig:methods}[A].
ProbCT processes multi-modal data. During inference,  its inputs comprise: image data denoted ${\bf y}$, acquired from $N^{\rm cam}$ viewpoints, each indexed $c$;  corresponding 3D camera locations $\{{\bf X}_{c} \}_{c=1}^{N^{\rm cam}}$; and the 3D coordinates ${\bf X}$ of a queried cloud location. 
In our implementation,  ${\bf X}$ is queried, if it passes a space carving cloud-mask, based on the multi-view images~\cite{levis2015airborne}.  
The ProbCT architecture is based on an encoder and a decoder. Here the encoder {\em increases} the dimension of the representation. 
The encoder is controlled by learned parameters 
${\boldsymbol \Theta}^{\rm enc}=[{\boldsymbol \Theta}^{\rm cam},{\boldsymbol \Theta}^{\rm domain},
{\boldsymbol \Theta}^{\rm image}]$, detailed below. Per ${\bf X}$, the encoder outputs a vector ${\bf u}({\bf X}|{\boldsymbol \Theta}^{\rm enc})$, whose dimensions are much larger than the combined dimensions of voxel and camera poses and the number of image pixels that relate to ${\bf X}$. 
A decoder ${\cal D}$ then acts on ${\bf u}$, {\em decreasing} dimensions down to a short, discrete representation of the function ${\hat P}_{{\bf X}}(\beta|{\bf y})$ at ${\bf X}$.  The decoder is controlled by learned parameters ${\boldsymbol \Theta}^{\rm dec}$.
Overall, the vector of system parameters is
\begin{equation}
    \label{eq:allparameters}   
    {\boldsymbol \Theta}=[{\boldsymbol \Theta}^{\rm enc}, {\boldsymbol \Theta}^{\rm dec}] \;.
\end{equation}
ProbCT thus preforms
\begin{equation}
     \label{eq:Dprob}
    {\hat P}_{{\bf X}}(\beta|{\bf y}, {\boldsymbol \Theta}) 
    = {\cal D} 
    \left[ 
       {\bf u}
          ({\bf X}|{\boldsymbol \Theta}^{\rm enc}),
          {\boldsymbol \Theta}^{\rm dec}
    \right] \;.
\end{equation}

The encoder includes parallel, independent parts. One part encodes the location  ${\bf X}_{c}$ of viewpoint $c$, yielding an encoded vector ${\bf g}^{\rm cam}({\bf X}_{c})$. This encoder part is a DNN having four fully-connected ReLU layers, each layer having 64 neurons.
The neuron weights constitute ${\boldsymbol \Theta}^{\rm cam}$.
Using the same  ${\boldsymbol \Theta}^{\rm cam}$, this encoder-part is applied to all $N^{\rm cam}$  locations, in parallel. A similar DNN structure whose neuron weights constitute ${\boldsymbol \Theta}^{\rm domain}$ encodes any location ${\bf X}$ of a queried voxel. This encoder part yields a vector ${\bf g}^{\rm domain}({\bf X})$.

Image content of relevance to ${\bf X}$ is encoded by the following steps, leading to a feature vector ${\bf v}({\bf X})$: \\
{\tt (1)}  {\em Extract a map of image features} across a wide field of view, irrespective of the 3D volumetric element ${\bf X}$, using a convolutional DNN, described below. The neuron weights constitute ${\boldsymbol \Theta}^{\rm image}$. The same DNN, using the same ${\boldsymbol \Theta}^{\rm image}$, operates in parallel on all $N^{\rm cam}$ images. To implement this architecture for fast run-time, the $N^{\rm cam}$ images are set in the batch dimension of the DNN input. \\ 
{\tt (2)} {\em Query} ${\bf X}$. The location ${\bf X}$ is projected to each camera $c$. 
This yields 
a set of {\em continuous-valued} image locations $\{ {\bf x}_c \}_{\forall c}$, which correspond to ${\bf X}$.  
{\tt (3)} {\em Sample image features} per ${\bf x}_c$.
The image feature map of step  {\tt (1)} is on a discrete (integer-valued) pixel grid. On the other hand, a projected location ${\bf x}_c$ by step  {\tt (2)} is continuous valued, hence generally at intermediate locations between image pixels. Thus, each image feature map is linearly interpolated and re-sampled at ${\bf x}_c$. \\
{\tt (4)} {\em Concatenate}  corresponding features $\forall c$ to a single vector ${\bf v}({\bf X})$.

A map of image features is derived by an off-the-shelf feature pyramid network (FPN)~\cite{lin2017feature}.
A pyramid suits the multi-scale nature of clouds.

The decoder ${\cal D}$ has nine fully-connected ReLU layers. The final layer of the ProbCT decoder {\em {outputs a vector}} of length $Q$, which is a discrete representation of ${\hat P}_{{\bf X}}(\beta|{\bf y}, {\boldsymbol \Theta})$. It corresponds to quantized values $\beta(q)=q\Delta\beta$, where
$q\in[0,\ldots ,Q-1]$ and $\Delta\beta$ is the quantization step of the extinction coefficient. 
The parameters $Q,\Delta\beta$ were set so as to  cover the entire range of $\beta$ in the training set, that is $Q\Delta\beta\ge \max \beta$. An ablation study for $\Delta\beta$ is {shown in the supplementary material}.
We set $Q=301,\Delta\beta=\unit[1]{\rm km^{-1}}$ in \cref{sec:sim}. In \cref{sec:airmspi},  $Q=101,\Delta\beta=\unit[0.5]{\rm km^{-1}}$.
This discretization is done only for the posterior probability  estimation; not during cloud rendering or error quantification.

\subsubsection{Inference}
\label{sec:infer}

The ProbCT model is based on a DNN, controlled by a set of parameters ${\boldsymbol \Theta}$. This set is learned by training, forming an optimal set of parameters ${\hat {\boldsymbol \Theta}}$. The trained system is then used to infer new, unknown scenes. Thus, during inference, we denote the output ${\hat P}_{{\bf X}}(\beta|{\bf y},{\hat {\boldsymbol \Theta}})$. 
Specifically, we estimate\footnote{In the {supplementary material} we show additional results using the estimated {\em posterior mean}. For the $\epsilon$ criterion in \cref{eq:erros}, the posterior mean performs somewhat worse to MAP by $\sim$3\%.} $\beta$ at ${\bf X}$ using the 
MAP criterion:
\begin{gather}
         {\hat \beta}({\bf X})={\hat q}({\bf X})\Delta\beta~~~~~~~~~~~~~~~~~~~~~~~~~~~~~~~~~~~~~~~~~~~~ \nonumber \\
     ~~{\rm where}~~
    {\hat q}({\bf X}) = 
    \arg\! \max_q 
    {\hat P}_{{\bf X}}(q\Delta\beta|{\bf y}, {\hat {\boldsymbol \Theta}}).
    \label{eq:q_hat}
\end{gather}
Using \cref{eq:q_hat} $\forall {\bf X}$ estimates the 3D volumetric object. 
Uncertainty can be quantified by the normalized entropy 
\begin{gather}
 H_{\bf X}^{\rm norm} = ~~~~~~~~~~~~~~~~~~~~~~~~~~~~~~~~~~~~~~~~~~~~~~~~~~~~~~\nonumber \\
    \frac{-\sum_{q=0}^{Q-1} 
       \left[
           {\hat P}_{\bf X}(q\Delta\beta|{\bf y},{\hat {\boldsymbol \Theta}})
           \log_2 {\hat P}_{\bf X}(q\Delta\beta|{\bf y},{\hat {\boldsymbol \Theta}})
       \right]}
       {\log_2 Q}
  \;.
  \label{eq:rel_entropy}
\end{gather}
Here $H_{\bf X}^{\rm norm}=0$ for absolute certainty, where ${\hat P}_{\bf X}(q\Delta\beta|{\bf y},{\hat{\boldsymbol \Theta}})$ is a delta function. On the other hand, $H_{\bf X}^{\rm norm}=1$ when ${\hat P}_{\bf X}(q\Delta\beta|{\bf y},{\hat{\boldsymbol \Theta}})$ is uniformly distributed, corresponding to maximum uncertainty.

\subsubsection{Training}
\label{sec:train}

A set of labeled pairs 
$\{ ({\boldsymbol \beta}^{\rm true}_n,{\bf y}_n)  \}_{n=1}^N$ 
is used for supervised training of ${\boldsymbol \Theta}$. 
During supervised training, the ProbCT model is exposed to four types of variables: $\bullet$ A large variety of objects (clouds). Thus, the model implicitly learns priors that express what is more probable or less probable to exist in a voxel of a cloud, in the context of other voxels. Hence the model learns  priors on the randomness of nature. $\bullet$ Multi-view images corresponding to each scene. Thus the model learns to relate objects to images, without solving or inverting RT. $\bullet$ Random samples of image noise, according to a physical noise model. Thus the ProbCT model implicitly learns uncertainty of object recovery relating to sensing noise. $\bullet$ Perturbations to the imaging geometry. Thus ProbCT learns to generalize CT in variable projections.  

There is a disadvantage for supervised training relying on  generated clouds. These clouds are based on types (see Sec.~\ref{sec:Synth}) emerging from pre-set conditions. Nature tends to be more complicated than anticipated. Often, clouds observed in the wild may deviate from these types (thus the trained distribution). To help ProbCT handle OOD scenes, supervised training is augmented by {\em self-supervised} training (see \cref{fig:ProbCT_train1}), relying on $M$ unlabeled scenes. For them, we only have acquired image data $\{{\bf y}_m\}_{m=1}^M$, but no corresponding volumetric data. 
 Such data partly corresponds to OOD clouds. While
${\boldsymbol \beta}^{\rm true}_m$ is unknown, we know the forward model (Eq.~\ref{eq:forwardF}) that converts an arbitrary ${\boldsymbol \beta}_m$ field 
to rendered multi-view images. The rendered images should have a good {\em appearance match} to 
the real acquired data ${\bf y}_m$. Hence, ${\boldsymbol \Theta}$ can be tuned, so that an appearance match measure is optimized. 
Such self-supervised learning uses only the physical forward model of RT, hence it is not sensitive to priors of cloud structure.

We now detail supervised training. For labeled data, the
true probability distribution at ${\bf X}$ is discretized and represented by a vector, whose $q^{\rm th}$ element is 
\begin{equation}
  P_{{\bf X}}^{\rm true}(q\Delta\beta)=  
      \begin{cases}
		    1 & \text{if}~~~ 
             q=\lfloor 
                 \beta^{\rm true}({\bf X})/\Delta\beta
                \rfloor\\
            0 & \text{otherwise}
		 \end{cases}\;.
  \label{eq:p_true}
\end{equation}
On the other hand, ProbCT infers a corresponding vector   
 ${\hat P}_{{\bf X}}(q\Delta\beta|{\bf y},{\boldsymbol \Theta})$. Training seeks to minimize the distance between these  discrete probability distributions. 
Distance between probability distributions is measured by the Kullback-Leibler (KL) divergence~\cite{joyce2011kullback}. The cross entropy of the distributions is
\begin{gather}
       {\rm CE}_{\bf X}({\bf y},{\boldsymbol \Theta})=
      {\rm CE}\{
       P_{{\bf X}}^{\rm true}(\beta), 
         {\hat P}_{\bf X}(\beta|{\bf y},{\boldsymbol \Theta})\} 
       ~~~~~~~~~~~~~~~~~~~ \notag \\
      ~~~~~~ =
       - \sum_{q}
      \left[P_{{\bf X}}^{\rm true}(q\Delta\beta) 
      \log {\hat P}_{\bf X}(q\Delta\beta|{\bf y},{\boldsymbol \Theta})
      \right] 
      ~~~~~~~\notag \\
       ~~~=
      -\log  {\hat P}_{{\bf X}}
       \left(
         \left.
           \left\lfloor 
               \frac{\beta^{\rm true}({\bf X})}{\Delta\beta}
            \right\rfloor
            \Delta\beta
         \right|
        {\bf y},{\boldsymbol \Theta}
      \right) 
     .~~
    \label{eq:ce}
\end{gather}
The last expression in (\ref{eq:ce}) is due to Eq.~(\ref{eq:p_true}).
The entropy~\cite{shannon2001mathematical} of the true distribution at ${\bf X}$ is
\begin{equation}
      H_{\bf X}^{\rm true}=
         H\{P_{{\bf X}}^{\rm true}(\beta)\}
         =0
         \;,
    \label{eq:H}
\end{equation}
due to Eq.~(\ref{eq:p_true}), independently of ${\boldsymbol \Theta}$. The KL divergence is 
\begin{equation}
      {\rm KL}
      \{ P_{{\bf X}}^{\rm true}(\beta),P_{\bf X}(\beta|{\bf y},{\boldsymbol \Theta})\} =
       H_{\bf X}^{\rm true} + 
       {\rm CE}_{\bf X}({\bf y},{\boldsymbol \Theta})
    \;.
    \label{eq:kl}
\end{equation}
Based on Eqs.~(\ref{eq:H},\ref{eq:kl}), the CE criterion (\ref{eq:ce}) is the key for optimization. 

Aggregating the CE over all voxels and labeled scenes, supervised training solves this optimization form
\begin{gather}
    (
         {\hat{\boldsymbol \Theta}}^{\rm enc}_{\rm super},
         {\hat{\boldsymbol \Theta}}^{\rm dec}_{\rm super}
    ) 
     = ~~~~~~~~~~~~~~~~~~~~~~~~~~~~~~~~~~~~~~~~~~~~~~~~~\nonumber\\
     \arg\min_{\boldsymbol \Theta}
    \sum_{n=1}^N 
       \sum_{{\bf X}} 
        w^{\rm cloud}_{{\bf X},n}
         {\rm CE}_{\bf X}
         ({\bf y}_n,
          {\boldsymbol \Theta})
     \;.
        \label{eq:OptiSuper}
\end{gather}
Here $w^{\rm cloud}_{{\bf X},n}$ is a weight. Most voxels in a scene are empty, having $\beta=0$. Just a small minority of voxels are in a cloud ($\beta>0$). We found in practice that if the contribution of empty voxels to the optimization loss (Eq.~\ref{eq:OptiSuper}) is not weighted down, the system trains to focus too much on void areas. Therefore, we weigh the loss.
{In all cases,  $w^{\rm cloud}_{{\bf X},n}=1$ if in scene $n$, $\beta^{\rm true}({\bf X})\geq\Delta\beta/2$, that is, $\bf X$ is a cloud voxel. A smaller weight is assigned to voxels that are empty ($\beta^{\rm true}({\bf X})<\Delta\beta/2$). Specifically,  $w^{\rm cloud}_{{\bf X},n}=0.01$ in \cref{sec:sim} and $w^{\rm cloud}_{{\bf X},n}=0.1$ in \cref{sec:airmspi}.}

Now, we detail the stage of self-supervised refinement using unlabeled data $\{  {\bf y}_m\}_{m=1}^M$ of $M$ imaged clouds. During self-supervised training, rendering ${\cal F}$ assumes a fixed default phase function and albedo by {setting the  droplet effective radius to $10\mu m$} and effective variance of 0.1, as in~\cite{levis2015airborne}.
Using ${\boldsymbol \Theta}^{\rm enc},
        {\boldsymbol \Theta}^{\rm dec}$,
running  \cref{eq:q_hat} $\forall {\bf X}$ estimates a 3D volumetric object denoted 
${\hat {\boldsymbol \beta}}_m({\bf y}_m,  
   {\hat {\boldsymbol \Theta}}^{\rm enc},
   {\hat {\boldsymbol \Theta}}^{\rm dec})$.
The forward model ${\cal F}$ of 3D RT  renders  images of 
${\hat {\boldsymbol \beta}}_m$.
Define a cost 
 \begin{gather}
      E({\boldsymbol \Theta}^{\rm enc},
        {\boldsymbol \Theta}^{\rm dec})
     =~~~~~~~~~~~~~~~~~~~~~~~~~~~~~~~~~~~~~~~~~~~~~~~~~~~~~~~\nonumber\\
     \sum_{m=1}^M
        \left\|
          {\bf y}_m -
          {\cal F} 
            \left\{ 
                {\hat {\boldsymbol \beta}}_m
                ({\bf y}_m,
                {\boldsymbol \Theta}^{\rm enc},
                {\boldsymbol \Theta}^{\rm dec})
            \right\} 
        \right\|_2^2.
  \label{eq:Em}
\end{gather}  
Then, learning is achieved by minimizing $E$. This minimization
leverages stochastic gradient descent based on {\em differential rendering}, that expresses how ${\cal F}$ changes by small deviations in ${\hat {\boldsymbol \beta}}_m$.
In principle, the whole set of parameters $({\boldsymbol \Theta}^{\rm enc},
                {\boldsymbol \Theta}^{\rm dec})$ can be optimized. However, we opted to keep the {\em encoder} parameters 
fixed at ${\hat{\boldsymbol \Theta}}^{\rm enc}_{\rm super}$ (obtained by Eq.~\ref{eq:OptiSuper}). The reasons are:
(i) The encoder learns general features of cloud images and viewing geometry, for use in scattering-based CT, insensitive to a specific cloud class. We wish to preserve the strong image features established  from  labeled data. 
(ii) The set size $M$ is small relative to $N$ because iterating the forward model is computationally expensive. Hence we refine only
${\boldsymbol \Theta}^{\rm dec}$:
\begin{equation}
    {\tilde{\boldsymbol \Theta}}^{\rm dec} = 
    \arg\min_{{\boldsymbol \Theta}^{\rm dec}} 
    E({\hat{\boldsymbol \Theta}}^{\rm enc}_{\rm super},
        {\boldsymbol \Theta}^{\rm dec})
    \;.
   \label{eq:image_cyclic}
\end{equation}
This optimization is initialized by 
${\hat{\boldsymbol \Theta}}^{\rm dec}_{\rm super}$
from Eq.~(\ref{eq:OptiSuper}). 


All operations required for \cref{eq:image_cyclic} are differentiable, except\footnote{Inference by \cref{eq:q_hat} is oblivious to differentiabilty.}  the  $\arg\!\max$ operator in \cref{eq:q_hat}.
We approximate this operator using a differential ${\tt Smoothmax}$ (Boltzmann)  operator~\cite{asadi2017alternative}.
Set a parameter $\alpha>0$. Define a normalized  amplified probability distribution
\begin{equation}
    { \Phi}_{{\bf X}}(q) = 
    \frac{ 
        \left[ {\hat P}_{{\bf X}}(q\Delta\beta|
            {\bf y},{{\boldsymbol \Theta}})
        \right]^\alpha }
        { \sum_{q} 
        \left[ 
            {\hat P}_{{\bf X}}(q\Delta\beta|
            {\bf y}, { {\boldsymbol \Theta}})
        \right]^\alpha}
    \;.
    \label{eq:amp_p}
\end{equation}
For  $\alpha\rightarrow\infty$, 
${ \Phi}_{{\bf X}}({\hat q})
  \rightarrow 
  \delta({q-{\hat q}})$ 
where ${\hat q}$ is given in \cref{eq:q_hat}.
We use $\alpha=10$.
Define \mbox{${\boldsymbol \Phi}_{{\bf X}}=[{ \Phi}_{{\bf X}}(0),{ \Phi}_{{\bf X}}(1),\dots,{ \Phi}_{{\bf X}}(Q-1)]^\top$} and \mbox{${\mathbf b}={\Delta\beta}\cdot[0,1,\dots,Q-1]$}, where $\top$ denotes transposition. 
A differential approximation to \cref{eq:q_hat},
yielding a continuous value
is
\begin{equation}
    {\hat \beta}({\bf X})\approx 
    {\mathbf b}{\boldsymbol \Phi}_{{\bf X}}\;.
    \label{eq:beta_diff}
\end{equation}
\cref{fig:methods}[B] demonstrates an example of the operation of \cref{eq:amp_p,eq:beta_diff}.

Supervised training is done by $\approx 100,000$ iterations of stochastic gradient descent via an Adam optimizer. An iteration uses 1000 randomly sampled query voxels. Supervised  and self-supervised  training use  learning rates of 5e-5 and 1e-5, respectively, a weight decay of 1e-5 and ran on a single NVIDIA GeForce RTX 3090 GPU.
For the results in \cref{sec:sim}, self-supervised training used 500 cloud scenes. Results in \cref{sec:airmspi} using real AirMSPI data performed self-supervision using three clouds.

\subsection{Synthetic 3D Clouds}
\label{sec:Synth}

Cloud simulations begin with environmental boundary conditions derived from real empirical data. For BOMEX, these are conditions over the Atlantic near Barbados~\cite{eytan2022shallow,siebesma2003large}. 
The Continental Active Surface-forced Shallow-cumulus (CASS) is a composite case for modeling shallow convection over land. CASS conditions follow 13 years of summertime observations by the Atmospheric Radiation Measurement Climate Facility of the US Department of Energy~\cite{zhang2017large} at Southern Great Plains (SGP).
Specifically, CASS represents a typical average behavior of selected 76 golden days having fair-weather shallow cumulus clouds. 
The HAWAII conditions correspond to warm maritime cumulus clouds: this model is initialized using a Hawaiian thermodynamic profile~\cite{heiblum2019core}, based on the 91285 PHTO Hilo radiosonde at 00Z on 21 August 2007.

Environmental conditions include aerosol concentrations. The aerosol concentration affects clouds in two main ways. First, aerosols affect feedbacks of cloud dynamics, consequently affecting the morphology of clouds and cloud fields. Second, a higher aerosol concentration leads to a higher concentration of cloud droplets, yet of smaller droplet size. This, in turn, affects the internal structure of the $\beta$ field in the cloud. Moreover, droplet size affects the scattering phase function and single-scattering albedo of the droplets, following Mie theory. These parameters 
then affect image formation, as described in Sec.~\ref{sec:3Drad}.  

The meteorological and aerosol conditions initialize a dynamical large eddy simulation (LES), which solves 
equations of atmospheric turbulence. LES is a key tool in cloud research~\cite{xue2006large}. We use a System for Atmospheric Modeling (SAM)~\cite{khairoutdinov2003cloud} for this part. The LES is coupled to a microphysical model that explicitly solves the governing process of droplet nucleation and growth (HUJI SBM~\cite{khain2004simulation}). The droplet size distribution is represented on a logarithmic axis sampled by 33 bins in the range [\unit[2]{$\mu$m}, \unit[3.2]{mm}].

\subsection{3D Radiative Transfer}
\label{sec:3Drad}

There are  $N$ generated volumetric cloud scenes, whose spatial distribution $\{{\boldsymbol \beta}^{\rm true}_n\}_{n=1}^N$ is known  in 3D. Each generated 3D cloud field (${\boldsymbol \beta}_n^{\rm true}$) yields multi-view images. The corresponding image data is denoted 
${\bf y}_n$, for $n=1\ldots N$. Rendering ground-truth images uses RT and an imaging noise model, both of which we describe. 
We first describe the process of 3D RT of incoherent light in heterogeneous media, illustrated in \cref{fig:methods}[C].  A 3D location is denoted $\bf X$.
Beside cloud droplets, air molecules affect RT.  Throughout this paper, we model the molecular extinction coefficient $\beta^{\rm air}({\bf X})$ using a summer mid-latitude vertical
distribution~\cite{levis2015airborne}, at altitudes in the range
$ [0,20]$km.
Atmospheric transmittance between any two points  ${\bf X}'$, ${\bf X}''$ is 
\begin{equation}
    \!\! \!\!  T({\bf X}', {\bf X}'') = 
    \exp{\left[ 
            -\!\!\int_{{\bf X}'}^{{\bf X}''} \!\!\!\!
            \{ 
            \beta({\bf X}) + \beta^{\rm air}({\bf X})
            \} d{\bf X}
        \right]}
     .
     \label{eq:transmittance}
\end{equation}
The medium 
is also characterized by a single-scattering albedo $\varpi({\bf X})$, and a scattering phase function. The phase function ${\tt p}({\bf X}, {\boldsymbol \omega} \cdot {\boldsymbol \omega}')$ expresses the relative portion of radiance scattered to the 3D direction unit-vector ${\boldsymbol \omega}$, when radiance is incident at ${\bf X}$ in direction ${\boldsymbol \omega}'$. The values of $\varpi$ and ${\tt p}$ stem from microphysical properties of the mixture~\cite{levis2017multiple} of particles in a voxel around ${\bf X}$, including air and water droplets.

At boundary point  ${\bf X}_0$ of the observed domain, the incident radiance $I({\bf X}_0, {\boldsymbol \omega})$ in direction ${\boldsymbol \omega}$ is known. 
Radiation  is then affected by the medium, generally multiple times, by interactions of scattering and absorption. As a result, a radiance 
field $I({\bf X}, {\boldsymbol \omega})$ encompasses the scene domain in all directions. 
This process is modelled by coupled and recursive 3D RT equations~\cite{chandrasekhar2013radiative}, sometimes referred to as {\em volume rendering} equations,
\begin{gather}
     I({\bf X}, {\boldsymbol \omega}) =I({\bf X}_0, {\boldsymbol \omega})T({\bf X}_0,{\bf X}) ~~~~~~~~~~~~~~~~~~~~~~~~~~~~ \notag \\
      +  \int_{{\bf X}_0}^{{\bf X}} 
     J({\bf X}',{\boldsymbol \omega})
     \left[\beta({\bf X}')+\beta^{\rm air}({\bf X}')\right]T({\bf X}',{\bf X}) d{\bf X}'
     \;,
     \label{eq:I}
\end{gather}
\begin{gather}
     \!\!\!\!J({\bf X}, {\boldsymbol \omega})\! = \frac{\varpi({\bf X})}{4\pi} \!\!  \int_{4\pi} 
     {\tt p}({\bf X}, {\boldsymbol \omega} \cdot {\boldsymbol \omega}') I({\bf X}, {\boldsymbol \omega}') d{\boldsymbol \omega}'
    \;.
     \label{eq:J}
\end{gather}

The radiance field is projected to $N^{\rm cam}$ observational cameras. 
Camera $c$ has a 3D center of projection at ${\bf X}_{c}$. In this camera, pixel $p$ corresponds to a line of sight having a particular direction, denoted ${\boldsymbol \omega}_{c,p}$. Projection of the scene to this pixel in this camera amounts to sampling the radiance field at  $I({\bf X}_{c}, {\boldsymbol \omega}_{c,p})$. 
Overall, the {\em forward model}   ${\cal F} \left({\boldsymbol \beta} \right)$ constitutes 3D RT followed by projection to all cameras, and consequent sampling to pixels. The acquired multi-view image data is represented by a vector 
\begin{equation}
     {\bf y}=
        {\cal N}
        \left\{
           {\cal F} \left({\boldsymbol \beta} \right)
        \right\}
             \;\;.
     \label{eq:forwardF}
\end{equation}
Here the operator ${\cal N}$ expresses introduction of random imaging noise. Noise specifications used in this paper are consistent with real instruments. They are described in Sec.~\ref{sec:noise}. 

In this paper, we use the SHDOM~\cite{levis2020git,loveridge2022git} RT solver. However, Monte-Carlo methods can also be used.
For the scattering phase function and the single-scattering albedo in AirMSPI rendering and training, we follow~\cite{levis2015airborne,levis2017multiple,ronen20214d,ronen2022variable} and use there 10$\mu {\rm m}$ droplet effective radius and effective variance of 0.1.

\subsection{Products}
\label{sec:product}

An estimate of ${\boldsymbol \beta}$ and its uncertainty propagate to the estimation of solar power generation. We now explain this conversion. Let
$I_{\lambda}({\bf X}, {\boldsymbol \omega}|{\boldsymbol \beta})$ be the radiance field of Eq.~(\ref{eq:I}), expressed as spectral radiance given ${\boldsymbol \beta}$, per wavelength $\lambda$.  The interaction of light with cloud droplets is rather insensitive to $\lambda$ in the visible and near-infrared spectral range $\Lambda$.
However, $I_{\lambda}({\bf X}, {\boldsymbol \omega}|{\boldsymbol \beta})$ depends on $\lambda$ due to scattering by air molecules. 
In the context of solar power generation, radiation power  is often quantified~\cite{duffie2013solar,xie2022fresnel} by the {\em global horizontal irradiance} (${\rm GHI}_\lambda$) in $[\frac{\rm W}{{\rm m}^2 {\rm nm}}]$, based on $I_{\lambda}({\bf X}, {\boldsymbol \omega}|{\boldsymbol \beta})$.

A PV panel has a transparent cover, eg., glass. Reflection by its surface decreases the cover transmittance  ${\tilde T}({\boldsymbol \omega})$ as a function of the direction of irradiance, ${\boldsymbol \omega}$.  Thus, we define and derive (see the Supplementary document) a corrected GHI that accounts for this effect:  
\begin{equation}
    {\widetilde {\rm GHI}}_{\lambda}({\bf X},{{\boldsymbol \beta}}) =  
    \int_{{\boldsymbol \chi} \cdot {\boldsymbol \omega}>0} 
    |{\boldsymbol \chi} \cdot {\boldsymbol \omega}| I_{\lambda}({\bf X}, {\boldsymbol \omega}|{\boldsymbol \beta})  
    {\tilde T}({\boldsymbol \omega}) 
   d {\boldsymbol \omega},
    \label{eq:GHI_tilda}
\end{equation}
where ${\boldsymbol \chi}$ is the nadir direction. 

A PV panel outputs a fixed voltage, while its current linearly depends on ${\widetilde {\rm GHI}}_\lambda$.
 The spectral response of the PV is ${\rm SR}_{\lambda}$ $[\frac{{\rm Amp} }{\rm W}]$.
 Then, 
\begin{equation}
     i_{\rm PV}({\boldsymbol \beta}) = 
     \int_{\Lambda} {\rm SR}_{\lambda} {\widetilde {\rm GHI}}_{\lambda}({\boldsymbol \beta}) d {\lambda} 
     ~~
     \left[
       \frac{\rm Amp}{{\rm m}^2}
     \right]
     \;.     
    \label{eq:PV_curr}
\end{equation}
Equation~(\ref{eq:PV_curr}) is then the basis for propagating the relative response to uncertainty in ${\hat {\boldsymbol \beta}}$ by
Equation~(\ref{eq:rel_responce}). 
Our results use ${\rm SR}_{\lambda}$ of typical monocrystalline Si solar panels~\cite{field1997solar}. 
The integral in Eq.~(\ref{eq:PV_curr}) is computed by sampling $\Lambda$ at 
[460,560,660,860,1060]nm, where each waveband is 20nm wide.

Precipitation is triggered~\cite{rosenfeld1994retrieving} when the droplet effective radius surpasses a critical value. At location ${\bf X}$, the droplet effective radius  is $r^{\rm e}_{\bf X}$. Let $\rho_w \approx 10^6{[\rm g/{\rm m}^3]}$ be the density of liquid water. The LWC and these variables are related~\cite{loeub2020monotonicity,marshak20053d} by
\begin{equation}
    r_{\bf X}^{\rm e}(\beta)=\frac{3{\mathcal{Q}}^{\rm eff}}{4\rho_w}\frac{{\rm LWC}({\bf X)}}{\beta({\bf X})}\;,
    \label{eq:r_e}
\end{equation}
where ${\mathcal{Q}}^{\rm eff}$ is the scattering efficiency of droplets, which is $\sim 2$ for visible light. In the core~\cite{acp-eshkol} of a cloud (See Sec.~\ref{sec:airmspi}), 
$    {\rm LWC}({\bf X})\approx{\rm LWC^{ad}}({Z})$. 
The function ${\rm LWC^{ad}}({Z})$ is computed~\cite{acp-eshkol}, given the cloud base altitude and the vertical temperature profile of the scene. These two parameters are obtained without requiring scattering CT: the cloud base is assessed by space-carving using the multi-view image data~\cite{levis2015airborne}. The temperature profile is sampled globally by various meteorological instruments~\cite{siebesma2003large}.  Ref.~\cite{acp-eshkol} uses typical environmental conditions over the Atlantic near Barbados, leading the function plotted in the supplementary material.
Overall, $r^{\rm e}_{\bf X}$ at the cloud core can be approximated by substituting  ${\rm LWC}({\bf X})$ by ${\rm LWC^{ad}}({Z})$ in \cref{eq:r_e}. We associate a voxel to the cloud core, if it is at least 100m away, horizontally, from the cloud edge. 
Let ${\bf X}\in {\cal Z}$ be the set of voxels at altitude $Z$, in the cloud core domain. Then, we set $r^{\rm e}$ per $Z$ using
\begin{equation}
    r^{\rm e}({\boldsymbol{\beta}})= \frac{1}{|{\cal Z}|}
    \sum_{{\bf X}\in {\cal Z}}
    r_{\bf X}^{\rm e}({{\beta}}_{\bf X})\;.
    \label{eq:r_e2}
\end{equation}

We also assess  $r_{\bf X}^{\rm e}$ in 3D without the adiabatic model, using the method of \cite{levis20213d}. Using the 3D fields $r_{\bf X}^{\rm e}$ and
 $\beta({\bf X})$,  \cref{eq:r_e} yields an estimate of ${\rm LWC}({\bf X})$, 
from which the AF is derived by Eq.~(\ref{eq:AF}).

\subsection{Distributions}
\label{sec:experdetails}

\subsubsection{Image noise}
\label{sec:noise}

The noise in AirMSPI training and real data complies with specifications described in~\cite{ASDC2013AirMSPI}.
For a formation of perspective cameras (as in CloudCT), we use noise specifications derived from the CMV4000 sensor, having a pixel size of $5.5 \times 5.5$ $\mu m^2$. The exposure time adjusts to the radiance that reaches the camera, so that the maximum image-pixel value corresponds to 90\% of the sensor full well, which is  13,500 photo-electrons. Thus, sampled radiance is converted to a Poissonian distributed photo-electron count. There are 13 photo-electrons per graylevel. The readout noise STD is 13 electrons. 
The camera uses 10bit quantization. 

\subsubsection{Probing a single voxel posterior}
\label{sec:singlevox}

In the case study illustrated in Fig.~\ref{fig:3d_bomex}[D], the dataset is made of clouds that have a variable voxel. The voxel has a random $\beta$, sampled from a bimodal probability
 \begin{equation}
 P^{\rm true}(\beta) = \begin{cases}
			\beta^{\rm low} & 
            {\rm  with\ probability\ 3/4} \\ 
            \beta^{\rm high} & \text{otherwise}
		 \end{cases}
   \;.
  \label{eq:posterior_toy}
\end{equation}
Here $\beta^{\rm low}, \beta^{\rm high}$ are normally distributed with expectations $42{{\rm km}^{-1}}$ and $75{{\rm km}^{-1}}$, respectively,  and have the same STD of  $\unit[5]{{\rm km}^{-1}}$.



\section{Code  availability}
\label{sec:codeavail}

The SAM code for generating cloud fields is available on the website \url{http://rossby.msrc.sunysb.edu/SAM.html}.

\appendix
\section*{Supplementary Material}

This is supplementary material for the main manuscript.

\section{Additional  results}
\subsection{BOMEX50 numerical  results}
In \cref{tab:bomex_results}, we detail numerical results for simulations described and plotted in the main manuscript for the BOMEX50 dataset. 
\begin{table}[t]
\caption{
A comparison based on the BOMEX50 dataset.
ProbCT  can lead to estimate $\hat \beta$ that achieves maximum a-posteriori (MAP) probability, or, alternatively, an expected (mean) value.
Physics-based~\cite{levis2015airborne} recovery time is about $\times1000$ longer than VIP-CT and ProbCT. We show the mean and standard deviation  of the $\epsilon,\delta$ criteria over all examples. ProbCT outperforms prior art. 
}
\centering
{
\begin{tabular}[t]{cccc}
\hline
\multicolumn{1}{c|}{Test} &
\multicolumn{1}{c|}{Method} &
 \multicolumn{1}{c|}{$\epsilon\% \downarrow$}  &  \multicolumn{1}{c}{$\delta\% \rightarrow0$}   \\ \hline
\multirow{3}{*}{{ {ID}}}& VIP-CT~\cite{ronen2022variable} & $31\pm10$ & $16\pm12 $ \\
                               &  ProbCT (MAP) & $\boldsymbol{ 30\pm10}$ & $15\pm12$ \\
                                &  ProbCT (mean)  & $31\pm11$ & $\boldsymbol{13\pm14}$ \\
\hline  
\multirow{3}{*}{{ {OOD}}}& VIP-CT~\cite{ronen2022variable} & $54\pm10$ & $12\pm13 $ \\
                               &  ProbCT (MAP) & $\boldsymbol{49\pm10}$ & $15\pm12$ \\
                               &  ProbCT (mean)  & $57\pm15$ & $\boldsymbol{-8\pm24}$ \\
\hline  
\hline  
&Physics-based~\cite{levis2015airborne}& $51\pm08$ & $31\pm 12$ 
\end{tabular}}
\label{tab:bomex_results}
\end{table}
Herein, performance is evaluated by these criteria
\begin{equation}
 \epsilon =\frac {\| { {\boldsymbol \beta}^{\rm true}} - {\hat {\boldsymbol \beta}} \|_1 }
 {\| { {\boldsymbol \beta}^{\rm true}}  \|_1}
  \;,\
   \delta =\frac {\| { {\boldsymbol \beta}^{\rm true}}\|_1 - \|{\hat {\boldsymbol \beta}} \|_1 }
 {\| { {\boldsymbol \beta}^{\rm true}}  \|_1}
  \;,
  \label{eq:erros2}
\end{equation}
where $0\leq \epsilon$ and $-1\leq \delta \leq 1$. Ideally, $\epsilon=0$ and $\delta=0$. 

Furthermore, we conducted ablation studies to evaluate how the quantization step $\Delta\beta$ affects the $\epsilon$ measure. The results are listed in the supplementary \cref{tab:delta_beta}.
\begin{table}[t]
\caption{
An ablation study of the quantization step $\Delta\beta$ in an ${\rm ID}$ test. A smaller $\Delta\beta$ requires a model of higher complexity and possibly more data to sufficiently train.  A good balance of accuracy vs. complexity is achieved at
\mbox{\unit[$\Delta\beta=1$]{${\rm km}^{-1}$}}.
}
\centering
{
\begin{tabular}[b]{ccccccc}
\hline
\multicolumn{1}{c|}{\unit[$\Delta\beta$]{${\rm km}^{-1}$}} &
\multicolumn{1}{c|}{0.1} &
\multicolumn{1}{c|}{0.5} &
 \multicolumn{1}{c|}{1} &
\multicolumn{1}{c|}{2} &
\multicolumn{1}{c}{10} 
 \\ \hline
\multicolumn{1}{c|}{$\epsilon\%$} &
\multicolumn{1}{c|}{\!\!$34\pm11$\!\!} &
\multicolumn{1}{c|}{\!\!$32\pm12$\!\!}  &
\multicolumn{1}{c|}{\!\!$33\pm11$\!\!}  &
\multicolumn{1}{c|}{\!\!$37\pm12$\!\!}  &
\multicolumn{1}{c}{\!\!$57\pm14$\!\!}  
\\
\hline  
\end{tabular}}
\label{tab:delta_beta}
\end{table}

\subsection{Interpretable example}
\label{sec:interp}

In this section, we design an interpretable example to assess  what ProbCT learns and infers.
The tests in the result section of the main manuscript and herein provide evidence supporting that ProbCT infers $P_{\bf X}(\beta|{\bf y})$, having similarity to a true $P^{\rm true}_{\bf X}(\beta|{\bf y})$ in limit and intermediate cases. 
\begin{figure}[t]
 \centering
  \includegraphics[width=1.0\linewidth]{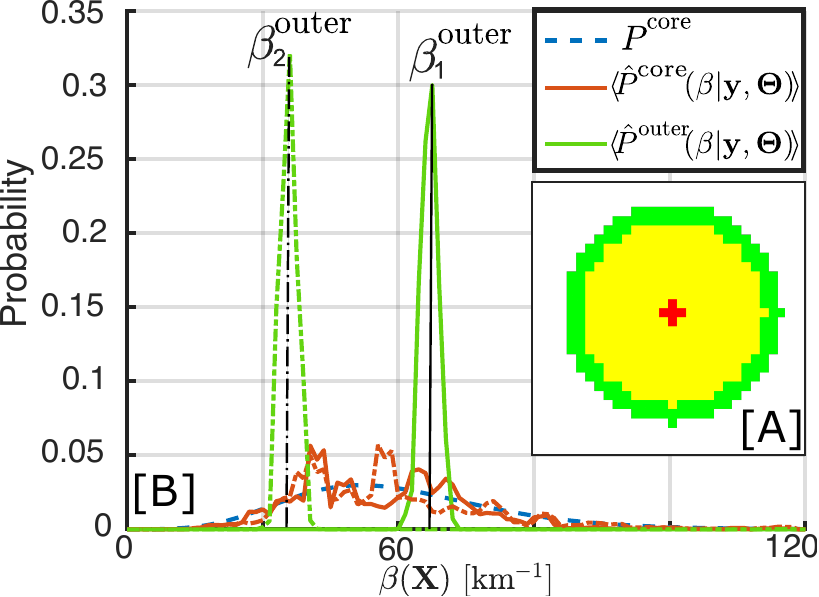}
  \caption{Red, yellow, and green stand for the cloud core, intermediate and outer shells, respectively.
  [A] A cross-section through the center of the spherical cloud.  [B] Plots
  for two different clouds in the test set, $\beta_1,\beta_2$, marked by solid and dash lines. They show
  the true (blue) probability distribution  of the core, estimated (red) probability distributions of the core,  and estimated outer-shell probability distributions (green).
  The actual sampled outer-shell values  are $\beta_1^{\rm outer},\beta_2^{\rm outer}$.
  }
  \label{fig:toy}
\end{figure}

Consider a spherical object (``spherical cloud'') having three concentric parts (\cref{fig:toy}A): A core having an unknown $\beta^{\rm core}$; an intermediate shell, whose optical thickness is known and very high; and an outer shell 
having an unknown $\beta^{\rm outer}$.
The clouds are observed from space by the geometry of the CloudCT formation. 
Voxels of the outer shell, mainly those on top, are directly exposed to light and to the cameras. Hence, measurements are sensitive to $\beta^{\rm outer}$. We thus expect  ${\hat \beta}$ in any outer voxel to be both accurate (close to  $\beta^{\rm outer}$) and having low uncertainty, being rather insensitive to the prior of the probability distribution of $\beta^{\rm outer}$, denoted
$P^{\rm outer}(\beta)$.

On the other hand, the middle of the sphere is {\em veiled} by the optically thick intermediate shell (a veiled core~\cite{forster2021toward,loveridge2023retrieving}). Light undergoes many scattering events until it reaches the core, and afterwards on the way to the cameras. The measurements in the cameras have noise, which overwhelms the core's signal. Hence, the measured signal is oblivious to $\beta^{\rm core}$. Therefore, we expect estimation of ${\hat \beta}$ in core voxels to be random, relying only on the prior probability distribution of $\beta^{\rm core}$, denoted $P^{\rm core}(\beta)$,
on which the system had trained. 

We generated 550 synthetic spherical clouds for training and 100 for testing. Each has spatially uniform shells with  
\begin{equation}
 {\beta}\left({\bf X}\right) =  
      \begin{cases}
			\beta^{\rm core} & ~~~~~~~~~~~
              \|{\bf X}-{\bf O}\|_2\leq \unit[60]{m} \\ 
            \beta^{\rm inter} & 
              ~\unit[60]{m}<\|{\bf X}-{\bf O}\|_2 \leq \unit[500]{m} \\ 
            \beta^{\rm outer} & 
             \unit[500]{m} < \|{\bf X}-{\bf O}\|_2\leq \unit[600]{m} \\
            0 & \text{otherwise}
		 \end{cases}
      \;,
  \label{eq:p_toy}
\end{equation}
where \mbox{${\bf O}=(0.8, 0.8, 1.28)$}km. In the intermediate shell,
$\beta^{\rm inter}=190{\rm km}^{-1}$ (visibility of $\approx 5$m).
In each cloud, the random values $\beta^{\rm core},\beta^{\rm outer}$ are drawn independently of each other from a log-normal probability
\begin{gather}
    P^{\rm outer}(\beta)=P^{\rm core}(\beta)=~~~~~~~~~~~~~~~~~~~~~~~~~~~~~~~~~~~~~~~~\nonumber\\
    ~~~~~~~~~~
    l(160/\beta)\exp{\{-8[\ln{(\beta/160)}+1]^2]\}}\;,
\end{gather}
where $l$ is a normalization constant and $\beta$ is in ${\rm km}^{-1}$, having expectation 
$\approx 61{\rm km}^{-1}$ and standard deviation 
$\approx 15{\rm km}^{-1}$.

After inference, let us empirically average the inferred probability distribution per shell. For example, let $|{\bf X}^{\rm core}|$ be the number of core voxels. The spatially-averaged inferred probability  is
\begin{equation}
   \langle {\hat P}^{\rm core}
     (\beta|{\bf y},{\boldsymbol \Theta})\rangle 
      =
      \frac{1}{|{\bf X}^{\rm core}|} 
      \sum_{{\bf X}^{\rm core}} 
              {\hat P}^{\rm core}_{\bf X}
              (\beta|{\bf y},{\boldsymbol \Theta}) \;.
 \label{eq:avergP}
\end{equation}
Supplementary \cref{fig:toy}B herein plots inferred probability distributions in two clouds. 
As expected (red lines), 
$\langle {\hat P}^{\rm core}
     (\beta|{\bf y},{\boldsymbol \Theta})
  \rangle \sim P^{\rm core}(\beta)$.
On the other hand, 
   $\langle {\hat P}^{\rm outer}
     (\beta|{\bf y},{\boldsymbol \Theta})\rangle$
is sharply peaked at the correct ground truth 
$\beta^{\rm outer}$, per cloud, with low uncertainty (green lines).

\section{Solar power calculations}
\label{sec:solar}
In this section, we detail the calculation of the photovoltaic (PV) current $i^{\rm PV}_{\bf X}({\boldsymbol \beta})$, due to solar energy 
(see the supplementary \cref{fig:PV_irradiance} herein).
\begin{figure}[t]
 \centering
\includegraphics[width=1.0\linewidth]{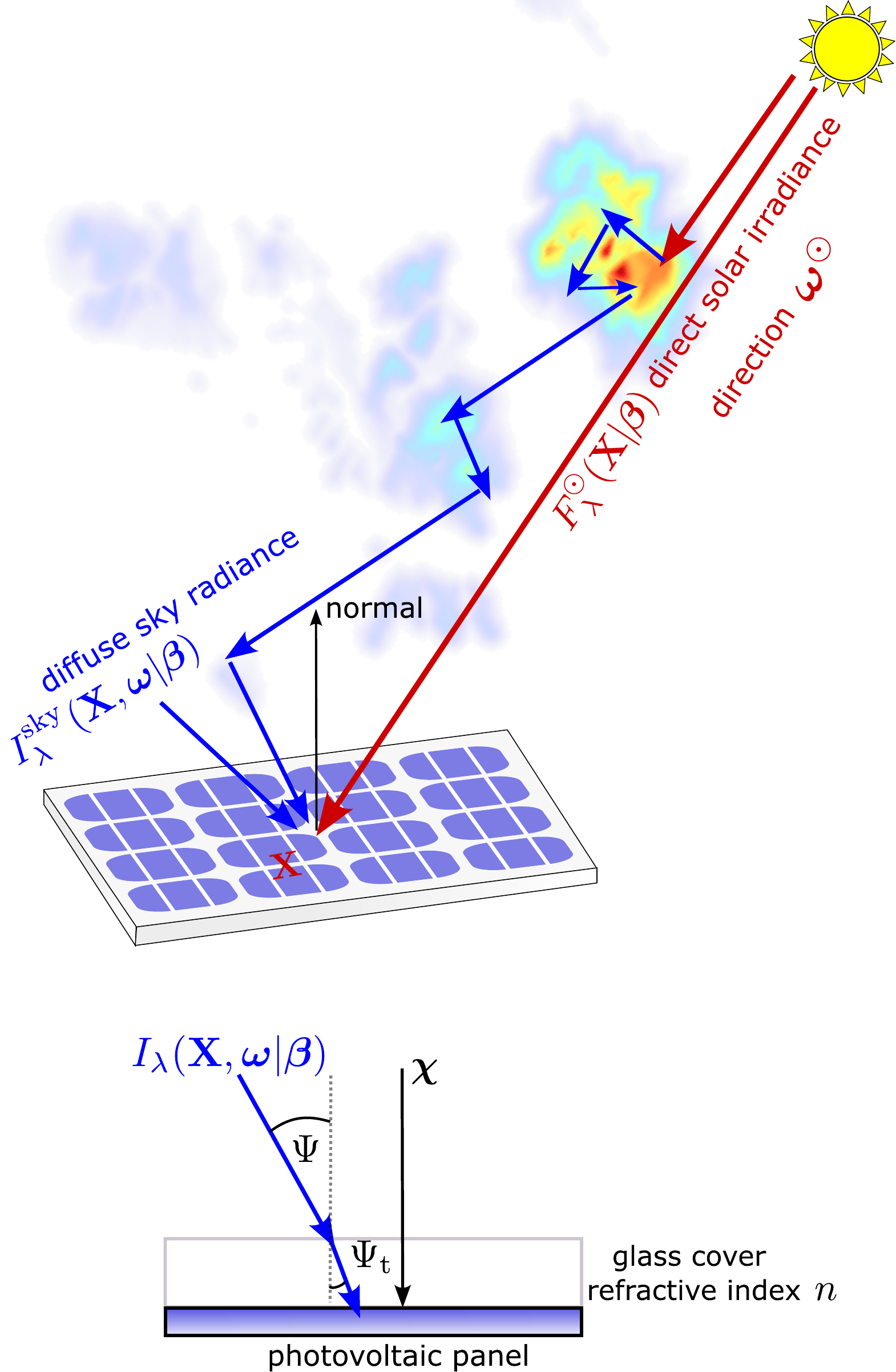}
  \caption{Direct solar irradiance and diffuse sky radiation. Both reach the PV panel and are influenced by the atmosphere. The panel cover changes the incident angle to a refracted angle. Reflection leads to  lower transmissivity by the cover.}
  \label{fig:PV_irradiance}
\end{figure}
The calculations include three elements: (a) The solar and sky irradiance that reaches the ground. (b) Transmission of incident irradiance through a PV cover material. (c) Conversion of radiation energy to electric current.

\subsection{Irradiance on the ground}
\label{sec:groundE}

Calculation of irradiance on the ground involves $\{i\}$ solar irradiance at the top of the atmosphere (TOA), $\{ii\}$ atmospheric extinction (including by clouds), leading to directly-transmitted solar irradiance, and $\{iii\}$  3D radiative transfer (RT) by the atmosphere, yielding diffuse sky irradiance.  The supplementary \cref{{fig:solar_and_pv_spectrum}} herein shows the experimentally measured solar irradiance~\cite{wehrli1985extraterrestrial} at the TOA. 
\begin{figure}[t]
 \centering
  \includegraphics[width=1.0\linewidth]{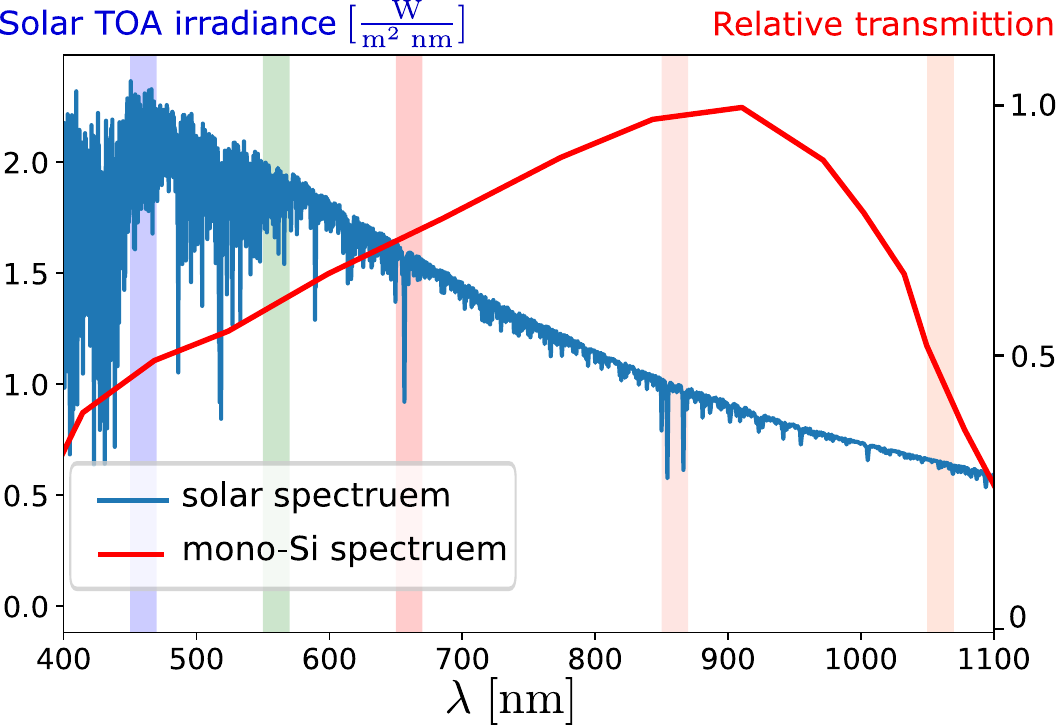}
  \caption{[Blue] Solar irradiance spectrum at the TOA. [Red] The spectral response of a monocrystalline Si solar PV panel. It is plotted here normalized by the maximum response, as the scale is factored out when calculating the relative response to uncertainty in ${\boldsymbol \beta}$.
  Five wavebands are marked, each having 20[nm] bandwidth. They sample the spectral functions.  
  }
  \label{fig:solar_and_pv_spectrum}
\end{figure}
Using the TOA irradiance, and given atmospheric content, RT calculations yield the radiance field 
$I_{\lambda}({\bf X}, {\boldsymbol \omega}|{\boldsymbol \beta})$, for any location ${\bf X}$ and direction ${\boldsymbol \omega}$, per wavelength $\lambda$.
The interaction of light with cloud droplets is relatively insensitive to $\lambda$ in the visible and near-infrared spectral range $\Lambda$. Therefore, we omit here the dependency of the cloud extinction coefficient ${\boldsymbol \beta}$ on $\lambda$. However, RT depends on $\lambda$ due to scattering by air molecules.

Let ${\boldsymbol \chi}$ denote the nadir direction.  For a location on the ground, only light coming from the upper hemisphere~\cite{marshak20053d} is relevant. Accordingly, the {\em global horizontal irradiance} (GHI)~\cite{duffie2013solar,xie2022fresnel} on the ground is 
\begin{equation}
   {\rm GHI}_{\lambda}({\bf X},{{\boldsymbol \beta}}) =
    \int_{{\boldsymbol \chi} \cdot {\boldsymbol \omega}>0} 
    |{\boldsymbol \chi} \cdot {\boldsymbol \omega}| I_{\lambda}({\bf X}, {\boldsymbol \omega}|{\boldsymbol \beta})
    \; d {\boldsymbol \omega}
\label{eq:GHI}
\end{equation}
in units of $[\frac{\rm W}{{\rm m}^2 {\rm nm}}]$.
While the ${\rm GHI}$ as given in the supplementary Eq.~(\ref{eq:GHI}) herein is a commonly used criterion  for solar power, it does not express the sensitivity of a PV panel to ${\boldsymbol \omega}$. Thus, we correct for this matter in the following.

\subsection{Transmission by a PV cover}
\label{sec:PVcover}

Consider a PV panel having a glass cover. For glass (neglecting its dispersion in this application),  the refractive index is  $n=1.5$ for $\lambda\in\Lambda$. Some of the incoming light is reflected by the cover, depending on 
${\boldsymbol \omega}$. Let
$\Psi({\boldsymbol \omega})=\arccos ( {\boldsymbol \chi} \cdot {\boldsymbol \omega})$ be the zenith angle of ${\boldsymbol \omega}$. Let the PV surface be horizontal. Light refracts into the cover at angle $\Psi_{\rm t}$ relative to ${\boldsymbol \chi}$, as illustrated in supplementary \cref{fig:PV_irradiance} herein. According to Snell’s law,
\begin{equation}
    \Psi_{\rm t}(\Psi)  = 
    \arcsin
    \left(
      \frac{1}{n}{\rm sin}{\Psi}    
      \right).
    \label{eq:snell}
\end{equation}
Based on Fresnel's equations~\cite{duffie2013solar,xie2022fresnel}, 
the transmissivity of the PV cover is 
\begin{equation}
    {\tilde T}({\boldsymbol \omega}) = 1 - 
    {\tilde R}[\Psi({\boldsymbol \omega})],
 \label{eq:Tpsi}
\end{equation}
where
\begin{equation}
    {\tilde R}[\Psi({\boldsymbol \omega})] = 
    \frac{{\rm sin}^2[{\Psi} - \Psi_{\rm t}(\Psi)]}
        {2{\rm sin}^2[{\Psi} + \Psi_{\rm t}(\Psi)]}  + 
    \frac{{\rm tan}^2[{\Psi} - \Psi_{\rm t}(\Psi)]}
         {2{\rm tan}^2[{\Psi} + \Psi_{\rm t}(\Psi)]}.
  \label{eq:Rpsi}
\end{equation}
We thus define a GHI that accounts for cover transmission
\begin{equation}
    {\widetilde {\rm GHI}}_{\lambda}({\bf X},{{\boldsymbol \beta}}) =  
    \int_{ {\boldsymbol \chi}\cdot {\boldsymbol \omega}>0} 
    |{\boldsymbol \chi} \cdot {\boldsymbol \omega}| I_{\lambda}({\bf X}, {\boldsymbol \omega}|{\boldsymbol \beta})  
    {\tilde T}({\boldsymbol \omega}) 
   d {\boldsymbol \omega} .
\label{eq:total_PV_flux}
\end{equation}

The reader can proceed now directly to the supplementary section~\ref{sec:current}, for conversion of this GHI to electric current. Meanwhile, we now provide implementation details. Direct solar irradiance arrives at direction 
${\boldsymbol \omega}^{\odot}$, having a corresponding zenith angle 
$\Psi({\boldsymbol \omega}^{\odot})$. This irradiance on the ground is denoted $F^{\odot}_{\lambda}({\bf X}|{\boldsymbol \beta})$, and accounts both for extinction by the atmosphere and the
$|{\boldsymbol \chi} \cdot {\boldsymbol \omega}^{\odot}|$ factor
that appears in supplementary Eq.~(\ref{eq:total_PV_flux}) herein.


Eq.~(\ref{eq:total_PV_flux}) herein can be divided to two components
\begin{align}
    {\widetilde {\rm GHI}}_{\lambda}({\bf X},{{\boldsymbol \beta}})  = &
    F_{\lambda}^{\odot}({\bf X}|{\boldsymbol \beta}) 
    {\tilde T}({\boldsymbol \omega}^{\odot})
    \\ \nonumber
     + 
    \int_{{\boldsymbol \chi} \cdot {\boldsymbol \omega}>0}
    &
    |{\boldsymbol \chi} \cdot {\boldsymbol \omega}| 
    I_{\lambda}^{\rm sky}({\bf X}, {\boldsymbol \omega}|{\boldsymbol \beta})  
    {\tilde T}({\boldsymbol \omega}) 
    \; d {\boldsymbol \omega},
\label{eq:total_PV_flux}
\end{align}
where $I_{\lambda}^{\rm sky}({\bf X}, {\boldsymbol \omega}|{\boldsymbol \beta})$ is the diffuse sky irradiance.
In our implementation, we obtain the fields $I_{\lambda}^{\rm sky}({\bf X}, {\boldsymbol \omega}|{\boldsymbol \beta})$ and $F_{\lambda}^{\odot}({\bf X}|{\boldsymbol \beta})$ using  the 
AT3D~\cite{loveridge2022git} code package. AT3D wraps a spherical harmonic discrete ordinate method (SHDOM) code of RT.

\subsection{Conversion to electric current}
\label{sec:current}

The 
 spectral response of monocrystalline silicon (mono-Si) PV~\cite{field1997solar} for $\lambda \in \Lambda$ is denoted by ${\rm SR}_{\lambda}$ $[\frac{{\rm Amp} }{\rm W}]$ and plotted in  the supplementary \cref{fig:solar_and_pv_spectrum} herein.
Then, the current generated per PV area is 
\begin{equation}
     i^{\rm PV}_{\bf X}({\boldsymbol \beta}) = 
     \int_{\Lambda} {\rm SR}_{\lambda} 
     {\widetilde {\rm GHI}}_{\lambda}({\bf X},{\boldsymbol \beta}) d {\lambda} 
     ~~
     \left[
       \frac{\rm Amp}{{\rm m}^2}
     \right]
     \;.     
    \label{eq:PV_curr2}
\end{equation}

\section{Liquid water content}
\label{sec:Precipitation}

In the core of a cloud, by the adiabatic model, the liquid water content (LWC) is only a function of the altitude $Z$~\cite{acp-eshkol} above the cloud base, that is,  
\begin{equation}
    {\rm LWC}({\bf X})\approx{\rm LWC^{ad}}({Z})\;.
    \label{eq:lwc_ad}
\end{equation}
The function ${\rm LWC^{ad}}({Z})$ can be calculated~\cite{acp-eshkol}. Such a function is shown in the supplementary \cref{fig:lwc_ad} herein.
\begin{figure}[t]
 \centering
  \includegraphics[width=1.0\linewidth]{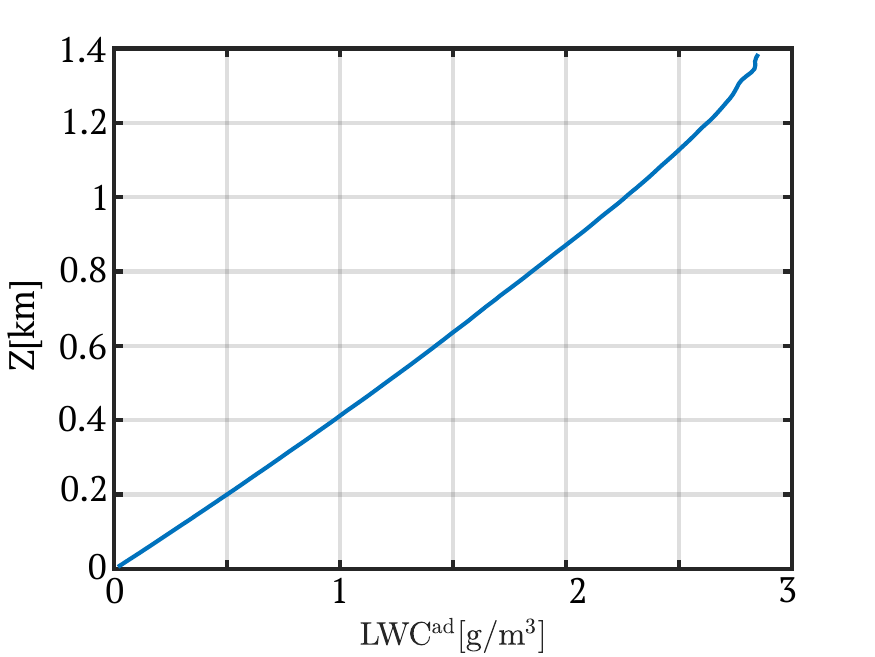}
  \caption{A plot of the cloud liquid water content (LWC) as a function of altitude $Z$ above the
  cloud base, according to the adiabatic model.
  }
  \label{fig:lwc_ad}
\end{figure}

\bibliography{sn-bibliography}

\end{document}